\documentclass{article}

\usepackage{PRIMEarxiv}

\usepackage[utf8]{inputenc} 
\usepackage[T1]{fontenc}    
\usepackage{hyperref}       
\usepackage{url}            
\usepackage{booktabs}       
\usepackage{amsfonts}       
\usepackage{nicefrac}       
\usepackage{microtype}      
\usepackage{lipsum}
\usepackage{fancyhdr}       
\usepackage{graphicx}       
\usepackage{amsmath}
\usepackage{subcaption}

\usepackage{float} 

\usepackage{algorithm}
\usepackage{algorithmic}

\graphicspath{{media/}}     

\title{exploring applications of topological data analysis in stock index movement prediction
}

\author{
  Dazhi Huang \\
  Guangdong University of Finance \\
  Guangzhou\\
  \texttt{211572421@m.gduf.edu.cn} \\
   \And
  Pengcheng Xu \thanks{\textit{Corresponding author}}\\
  Guangdong University of Finance \\
  Guangzhou \\
  \texttt{47-105@gduf.edu.cn} \\ \\
  \AND
  Xiaocheng Huang \\
  Guangdong University of Finance \\
  Guangzhou\\
  \texttt{221617231@m.gduf.edu.cn} \\
  \And
  Jiayi Chen \\
  Guangdong University of Finance \\
  Guangzhou \\
  \texttt{22154B12@m.gduf.edu.cn} \\
}

\begin{document}
\maketitle

\begin{abstract}
Topological Data Analysis (TDA) has recently gained significant attention in the field of financial prediction. However, the choice of point cloud construction methods, topological feature representations, and classification models has a substantial impact on prediction results. This paper addresses the classification problem of stock index movement. First, we construct point clouds for stock indices using three different methods. Next, we apply TDA to extract topological structures from the point clouds. Four distinct topological features are computed to represent the patterns in the data, and 15 combinations of these features are enumerated and input into six different machine learning models. We evaluate the predictive performance of various TDA configurations by conducting index movement classification tasks on datasets such as CSI, DAX, HSI and FTSE providing insights into the efficiency of different TDA setups.

\end{abstract}


\section{Introduction}
In the field of quantitative investment, classifying stock index movements is crucial for constructing investment strategies and managing risk. The intrinsic information embedded within stock indices is of great significance but is challenging to extract. A considerable amount of work has been done to explore stock index data and forecasting through various methods, including statistical approaches and machine learning techniques [\cite{bustos_stock_2020}, \cite{chu_fuzzy_2009}, \cite{liao_forecasting_2010}, \cite{lu_efficient_2011}, \cite{roh_forecasting_2007}, \cite{wang_forecasting_2011}, \cite{leung_forecasting_2000}]. While these methods successfully capture the main periodicity and volatility of stock index movements, they often overlook the underlying topological structure present within the data. In this context, Topological Data Analysis (TDA) combined with machine learning offers a promising approach for classifying stock index movements. However, the vast amount of available data in the stock market and the variety of topological features that can be utilized—along with the preference of different machine learning models for different data types and features—make it challenging to identify the optimal combination. These challenges have inspired us to explore the most effective and efficient way to apply TDA in the stock market.

TDA is a mathematical framework that uses concepts and methods from topology to analyze and express complex large datasets. It treats datasets as point clouds. Differing from traditional data analysis that focuses on individual points, TDA emphasizes the analysis of the topological shape and overall structure of point clouds, outputting topological features that reflect the organizational patterns and inherent relationships within the datasets. The foundation of TDA is developed by Herbert Edelsbrunner, Afra Zomorodian, Gunnar Carlsson  and several other people (see \cite{edelsbrunner_topological_2002},\cite{zomorodian_computing_2004},\cite{carlsson_topology_2009}).

\subsection{Previous work}

To integrate Topological Data Analysis with machine learning, various vectorized topological features have been introduced. A k-dimensional Betti number counts the number of k-dimensional holes and identifies which topological features persist. Umeda\cite{umeda_time_2017} introduced the Betti curve (Betti sequence) as a tool to enhance time series analysis in conjunction with machine learning. Their empirical results demonstrate the robustness of the Betti curve in both chaotic and non-chaotic time series classification tasks.

Goel et al.\cite{goel_topological_2020} applied Takens embedding theorem, utilizing the time-delay technique on stock data. Based on this embedding, they constructed the Vietoris–Rips complex. Subsequently, the authors computed the \( L^p \)-norm of the persistence landscape and integrated it with Enhanced Indexing (EI) to identify the most profitable portfolio. Their empirical results indicate that the combination of TDA and the \( L^p \)-norm of the persistence landscape outperforms traditional statistical models used in Enhanced Indexing (EI). Similarly, Yen and Cheong \cite{yen_using_2021} examined the correlations between individual stocks in the Singapore and Taiwan stock markets. These correlations were then transformed into distances, which were used to construct TDA features aimed at evaluating the overall market risk. Their results identified consistent topological signatures, such as Betti numbers, Euler characteristics, and persistent entropy, that were associated with market crashes. These findings suggest that such topological features could serve as potential early indicators of impending market crashes. Majumdar and Laha \cite{majumdar_clustering_2020} developed RF-TDA, a method based on Topological Data Analysis (TDA), to classify time series data in the financial market. They applied RF-TDA to real-world stock price data for financial time series classification tasks. Their results suggested that the topological features of stock price time series are distinct and can be effectively identified using TDA. These studies demonstrate that TDA can efficiently analyze financial data.
\subsection{Contribution}
This paper investigates the integration of Topological Data Analysis (TDA) with machine learning (TDA-ML) for financial market analysis. Our primary contribution lies in systematically examining the effects of various TDA feature combinations and methods for constructing point clouds to determine the optimal TDA-ML configuration. We propose a comprehensive methodology for analyzing financial data through TDA, applying it to real-world stock market data. Specifically, we explore the performance of TDA-ML by constructing three categories of point clouds and evaluating 15 feature combinations with six different machine learning models. Through extensive experimentation, we identify the most effective configuration for stock market prediction.Additionally, we provide a new perspective for investigating the topological properties of index constituent stock data.

\subsection{Outline}

In Section \ref{Persistent homology and TDA features}, we explain some basic concepts of TDA, including persistent homology and TDA features. Section \ref{methodology} outlines the main methodology and workflow of our work: We start with the datasets to construct point clouds, presenting three different methods for doing so. Although there are various rules for generating complexes from point clouds, such as VR complexes, \v{C}ech complexes, and alpha complexes (see Edelsbrunner and Harer \cite{edelsbrunner2022computational}), we have decided to use only the alpha complex after experimental comparisons. Once the complex is determined, its persistent homology groups can be smoothly defined and calculated, leading to the generation of persistence diagrams and various topological features. We select the Betti sequence, persistence landscape, and persistent entropy as the three topological features for subsequent machine learning input.   Section \ref{Experiments} describes the experimental setup, the data used, and presents the main empirical results of our experiments. Finally, Section \ref{conclusion} concludes the paper by summarizing our empirical findings.

\section{Persistent Homology and TDA Features}\label{Persistent homology and TDA features}
Complexes and persistent homology are key concepts of TDA. Since they can be found in many existing references, we provide only a brief introduction in Section \ref{sec_per_hom}. For more details, we refer Edelsbrunner and Harer's book \cite{edelsbrunner2022computational}. We also present several topological features associated to the output of persistent homology in Section \ref{sec_top_fea}, which will be used as inputs for the machine learning models.

\subsection{Persistent Homology} \label{sec_per_hom}

In topology, we can compute the homology groups for each complex. Given a point cloud, as we build the complex and increase the number of cells, we obtain different complexes, and their corresponding homology groups will change. Persistent homology is the tool that records these key changes. By constructing a filtration process for a point cloud, it tracks the topological features (such as connected components, loops, and voids) as they "appear" (birth) and "disappear" (death) at various scales. These features are visually represented using barcodes or persistence diagrams, providing an intuitive view of their persistence. This method helps identify both global and local geometric characteristics of data, making it a powerful tool for analyzing complex datasets, denoising, and feature extraction. It has been widely applied in fields such as bioinformatics, image processing, and time series analysis.

\begin{figure}[H]
\centering
\includegraphics[width=0.8\textwidth,height=.15\textwidth]{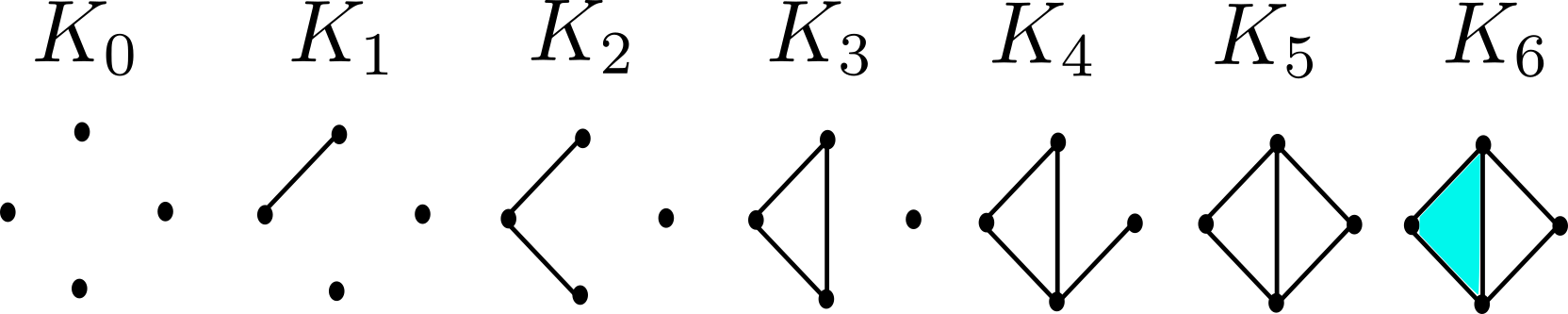}
\caption{A filtration of a complex from a point cloud of four points.}
\label{complex1}
\end{figure}

\begin{center}
\begin{tabular}{|c|c|c|c|c|c|c|c|}
  \hline
  $H_p$ & $K_0$ & $K_1$ & $K_2$ & $K_3$ & $K_4$ & $K_5$ & $K_6$   \\ \hline
  0 & $\mathbb{Z}^4$ & $\mathbb{Z}^3$ & $\mathbb{Z}^2$ & $\mathbb{Z}^2$ & $\mathbb{Z}$ & $\mathbb{Z}$ & $\mathbb{Z}$  \\
  1 & 0 & 0 & 0 & $\mathbb{Z}$ & $\mathbb{Z}$ & $\mathbb{Z}^2$  & $\mathbb{Z}$   \\ \hline
  Betti(p) & $K_0$ & $K_1$ & $K_2$ & $K_3$ & $K_4$ & $K_5$ & $K_6$   \\ \hline
  0 & 4 & 3 & 2 & 2 & 1 & 1 & 1  \\
  1 & 0 & 0 & 0 & 1 & 1 & 2 & 1 \\
  \hline
\end{tabular}
\end{center}

Consider a point cloud with only four vertices, as shown in Figure \ref{complex1}. As the radius parameter varies, an increasing number of cells are added to the complex, resulting in a sequence of complexes known as a filtration. At each step, we obtain a $K_i$, from which we can compute all its p-dimensional homology groups and Betti numbers. The newly added cells $\alpha^p$ (where k is the dimension) in each $K_i$ will cause the complex to undergo one of the following two changes: (1) form a new p-dimensional hole, thereby increasing the p-dimensional Betti number by one, or (2) form a pairing with a cell of one lower dimension $\beta^{p-1}$, thereby decreasing the (p-1)-dimensional Betti number by one. For each $i<j$, we have the inclusion map from $K_i$ to $K_j$, which induces a homomorphism between the p-dimensional homology groups: $f_{p}^{i,j}:H_p(K_i)\to H_p(K_j).$ The p-dimensional (i,j) persistent homology group is defined as the image of $f_{p}^{i,j}$, that is, $H_p^{i,j}=Im f_{p}^{i,j}.$

Persistent homology offers a multi-scale perspective on the data, allowing for the tracking of various qualitative features across different resolution scales. This culminates in a topological signature known as a persistence diagram. While persistence diagrams hold potentially valuable information about the underlying data set, they form a multi-set that, when equipped with the Wasserstein distance, constitutes an incomplete metric space. As a result, this renders them unsuitable for the application of statistical and machine learning tools aimed at processing topological features(see Goel et al.\cite{goel_topological_2020}).

\begin{figure}[H]
\centering
\includegraphics[width=0.8\textwidth,height=.4\textwidth]{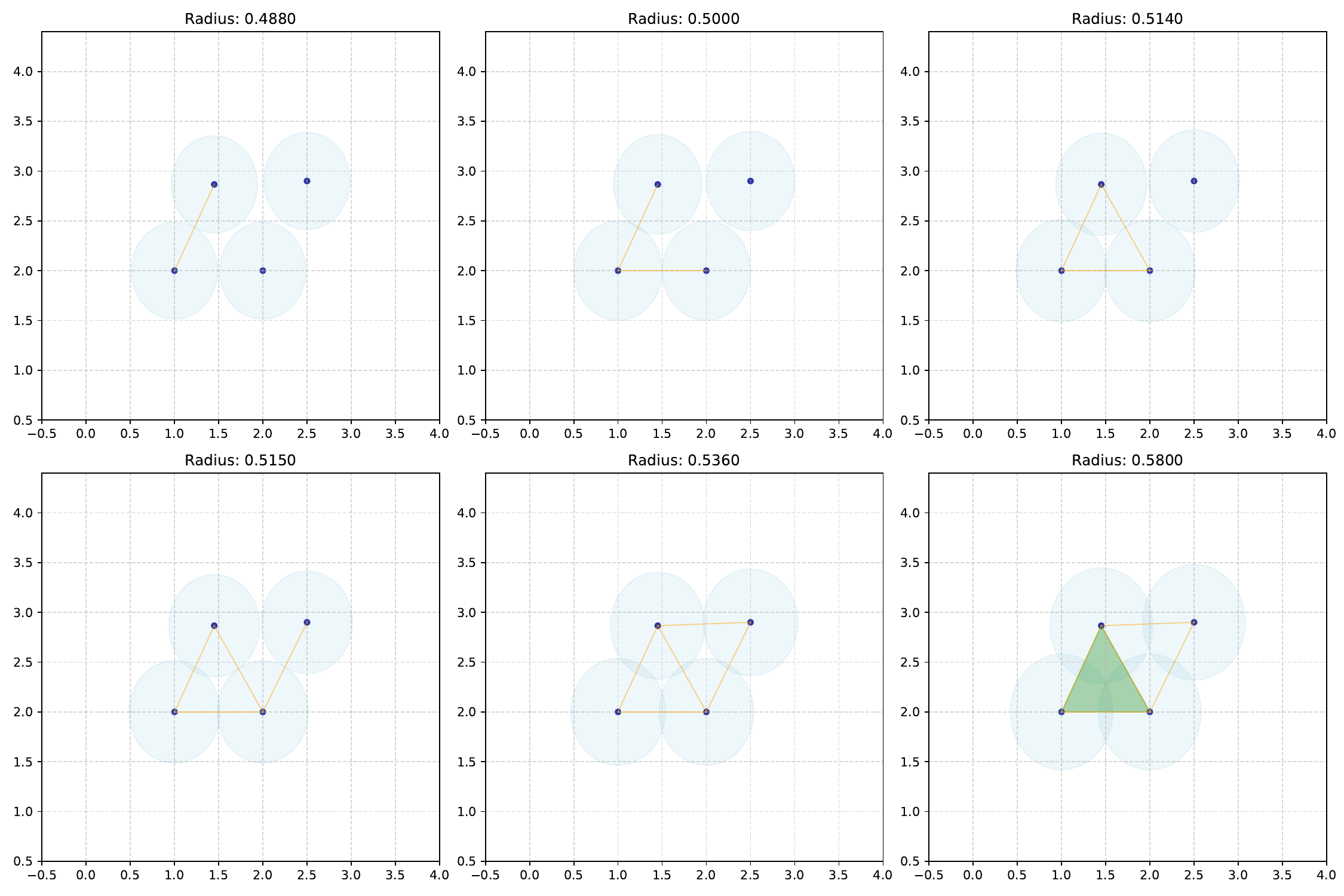}
\caption{The geometric realization of \v{C}ech complex of Figure \ref{complex1}.}
\label{vertex}
\end{figure}

\begin{figure}[H]
\centering
\includegraphics[width=0.4\textwidth,height=.3\textwidth]{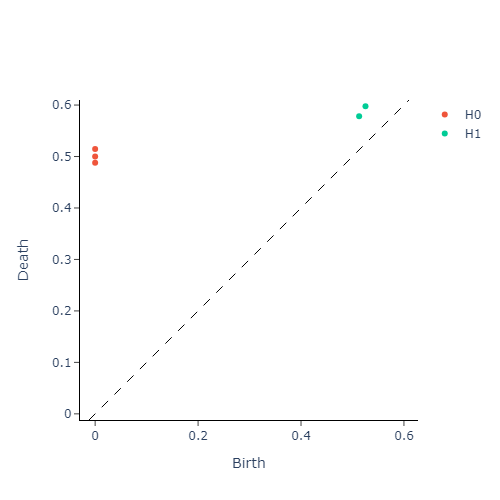}
\caption{The corresponding persistence diagram of Figure \ref{complex1}.}
\label{pd}
\end{figure}

The persistence diagram is the output of persistent homology. It consists of vectors, each of which is three-dimensional, containing the dimension of a cell in the complex, as well as its birth and death times. We illustrate its persistence diagram with the complex in Figure \ref{complex1} as an example. In Figure \ref{vertex}, we realize the filtration with \v{C}ech complex, with radii specified for each subgraph. Figure \ref{pd} shows the birth-death information of each cell in Figure \ref{vertex}.

\subsection{Topological Features}\label{sec_top_fea}
\subsubsection{Betti Curve}
The Betti curve is a vectorized feature derived from persistent homology, particularly suitable as input for machine learning models. It captures essential characteristics of data by analyzing its topological structure, effectively describing the shape and organization of datasets (see Umeda\cite{umeda_time_2017}).

\begin{figure}[H]

    \centering
    \begin{minipage}[c]{0.48\textwidth}
        \centering
        \includegraphics[height=0.2\textheight]{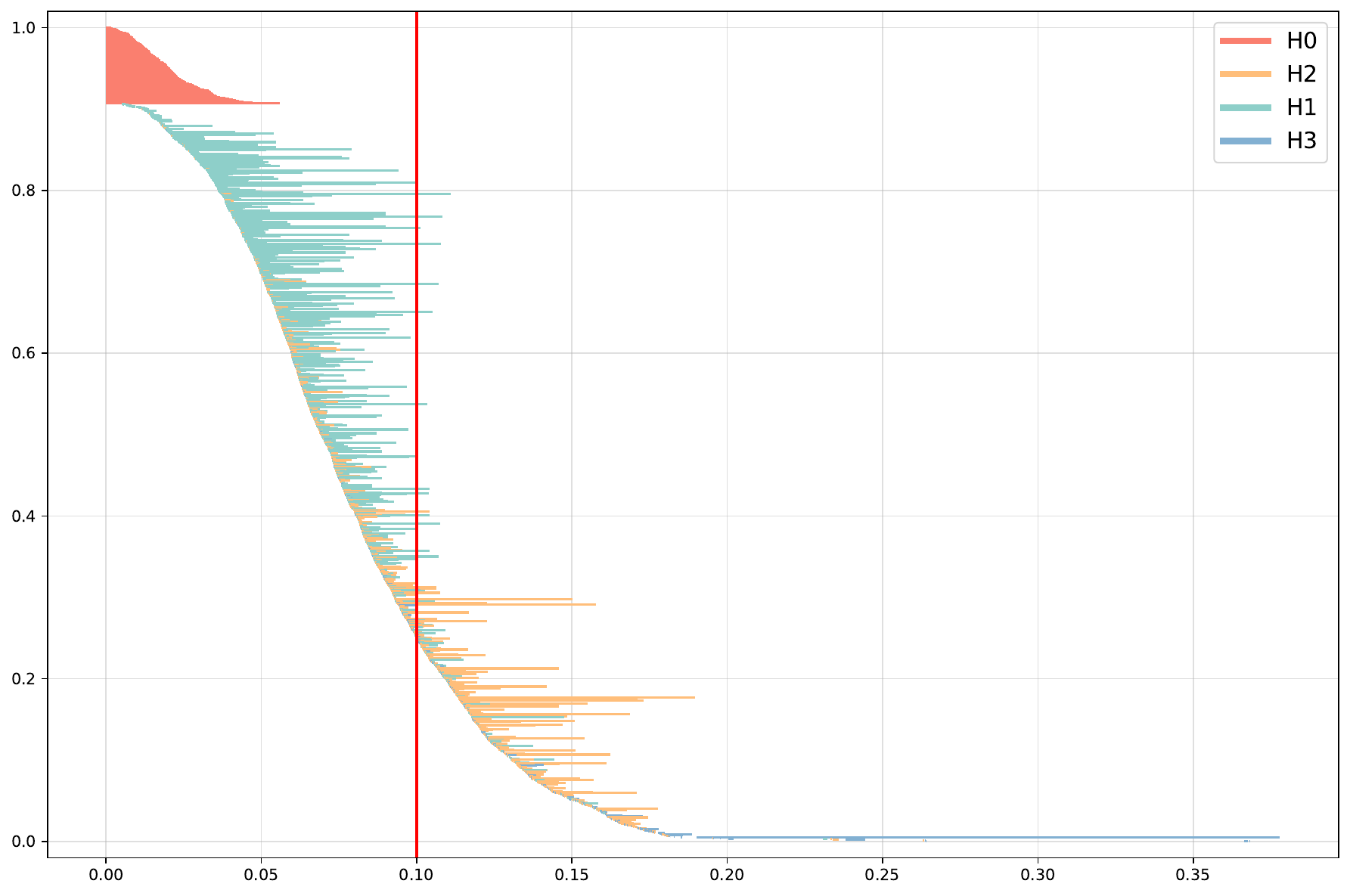}
        \subcaption{Persistence barcodes}
        \label{fig:BarCode}
    \end{minipage}
    \begin{minipage}[c]{0.48\textwidth}
        \centering
        \includegraphics[height=0.2\textheight]{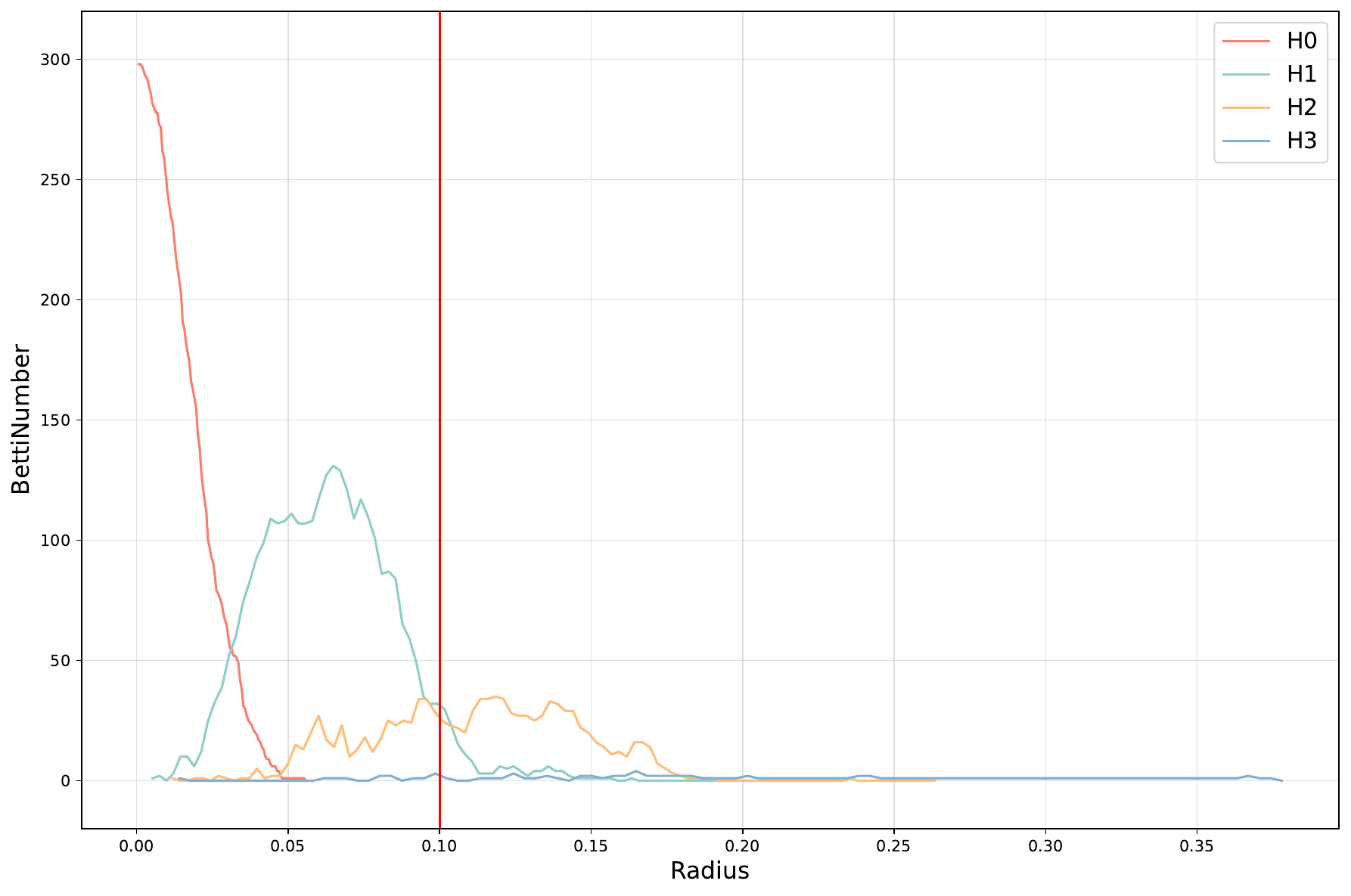}
        \subcaption{Betti curve}
        \label{fig:BettiCurve}
    \end{minipage}
    \caption{Illustration of the transformation from persistence barcodes to the Betti curve. The vertical red line marks a consistent radius across both subfigures. (\subref{fig:BarCode}) shows the persistence barcodes of the CSI300 Index on June 29, 2020, highlighting birth-death pairs of topological features as horizontal lines. (\subref{fig:BettiCurve}) shows the corresponding Betti curve derived from these barcodes.}
    \label{bcAndPb}

\end{figure}

Let \(BN_d(r)\) denote the \(d\)-dimensional Betti number of \(\mathbb{X}\) at a filter radius \(r\). To produce a Betti curve vector of finite length, we restrict the radius parameter to \(0 < r < E\), where \(E\) is set as the time at which the last barcode in the given dimension \(d\) vanishes. The \(i\)-th element of the \(d\)-dimensional Betti curve vector is given by \(BS_d(i) = BN_d(i \cdot E / m_d)\), where \(m_d\) represents the designated vector size. In this paper, \(m_d\) is set to 100. Figure \ref{bcAndPb} demonstrates the conversion process from persistence barcodes to the corresponding Betti curve.

\subsubsection{Persistent Entropy}
Persistent entropy, introduced by Chintakunta et al.\cite{chintakunta_entropy-based_2015}, quantifies the probability distribution of the lengths of persistent barcodes by calculating the Shannon entropy of each barcode (see Myers et al.\cite{myers_persistent_2019}). A high persistent entropy value indicates a complex topological structure in the raw data, while a low value corresponds to a simplified topological structure. Mathematically, the persistent entropy \( E(F_d) \) is defined as follows:
\begin{equation}
    E(F_d) = -\sum_{j \in J_d} p_j \cdot \log(p_j)
\end{equation}
where \( J_d \) is the collection of all persistent barcodes in dimension \( d \). For each birth-death pair (barcode) \( (a_j, b_j) \in J_d \), the corresponding lifetime is \( l_j = b_j - a_j \). The total lifetime of the barcode collection in dimension \( d \) is given by \( L_d = \sum_{j \in J_d} l_j \), and the probability associated with barcode \( j \) is \( p_j = \frac{l_j}{L_d} \).

\subsubsection{Total Persistence}
Specifically, in this paper, we select \( \sum_{j \in J_d} l_j \cdot \log(l_j) \) as the total persistence feature, which assesses the cumulative lifetime differences across data. During our experiment, We  unexpectedly discovered that it perform well in classification model.

\subsubsection{Persistence Landscape}

 To effectively embed persistence diagrams into a Hilbert space for time series analysis, the persistence landscape introduced by Bubenik \cite{bubenik2015statistical} is employed. This method involves a sequence of real-valued functions that encapsulate the essential information contained in persistence diagrams. In comparison to persistence diagrams, persistence landscapes offer several benefits, including their structure within Hilbert space and their stability(see Bubenik \cite{bubenik2015statistical}, Chazal et al.\cite{chazal2015subsampling}, Fasy et al.\cite{fasy2018robust}).

\begin{algorithm}[H]
\caption{Obtaining a Persistence Landscape}
\begin{algorithmic}[1]
\label{algo:landscape}
\STATE \textbf{Step 1:} Rotate the persistence diagram clockwise by \(45^\circ\).
\STATE \textbf{Step 2:} For each homology feature, treat it as a right angle vertex.
\STATE \textbf{Step 3:} Draw isosceles right triangles from each homology feature.
\STATE \textbf{Step 4:} Collect all the triangles formed.
\FOR{each integer \(m\) in the collection of triangles}
    \STATE \textbf{Step 5:} Obtain the \(m\)-th landscape function as the point-wise \(m\)-th maximum of all triangles drawn.
\ENDFOR
\end{algorithmic}
\end{algorithm}

\begin{figure}[H]
    \centering
    \begin{minipage}[c]{0.48\textwidth}
        \centering
        \includegraphics[height=0.2\textheight]{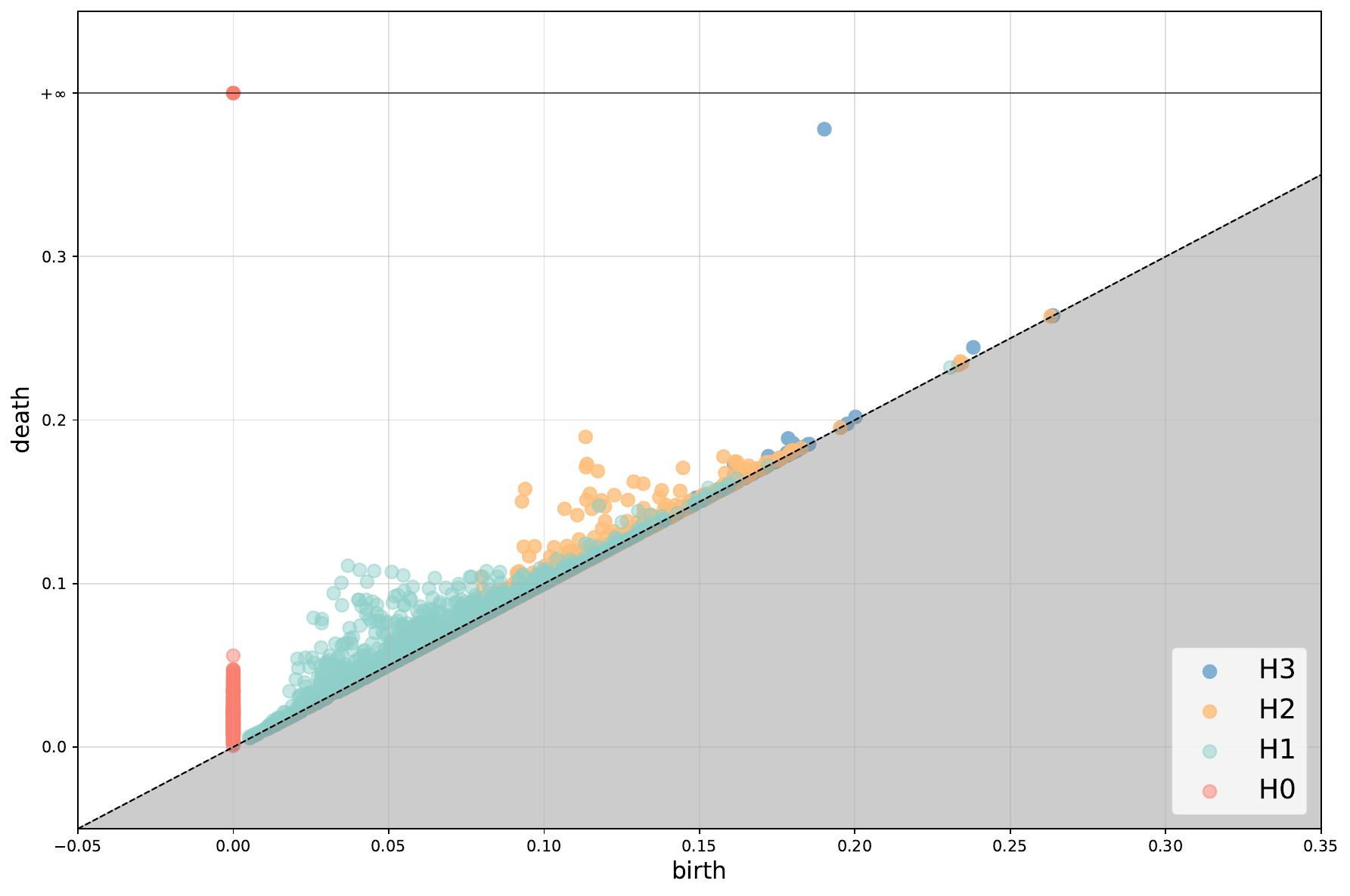}
        \subcaption{Persistence diagram}
        \label{fig:PD1}
    \end{minipage}
    \begin{minipage}[c]{0.48\textwidth}
        \centering
        \includegraphics[height=0.2\textheight]{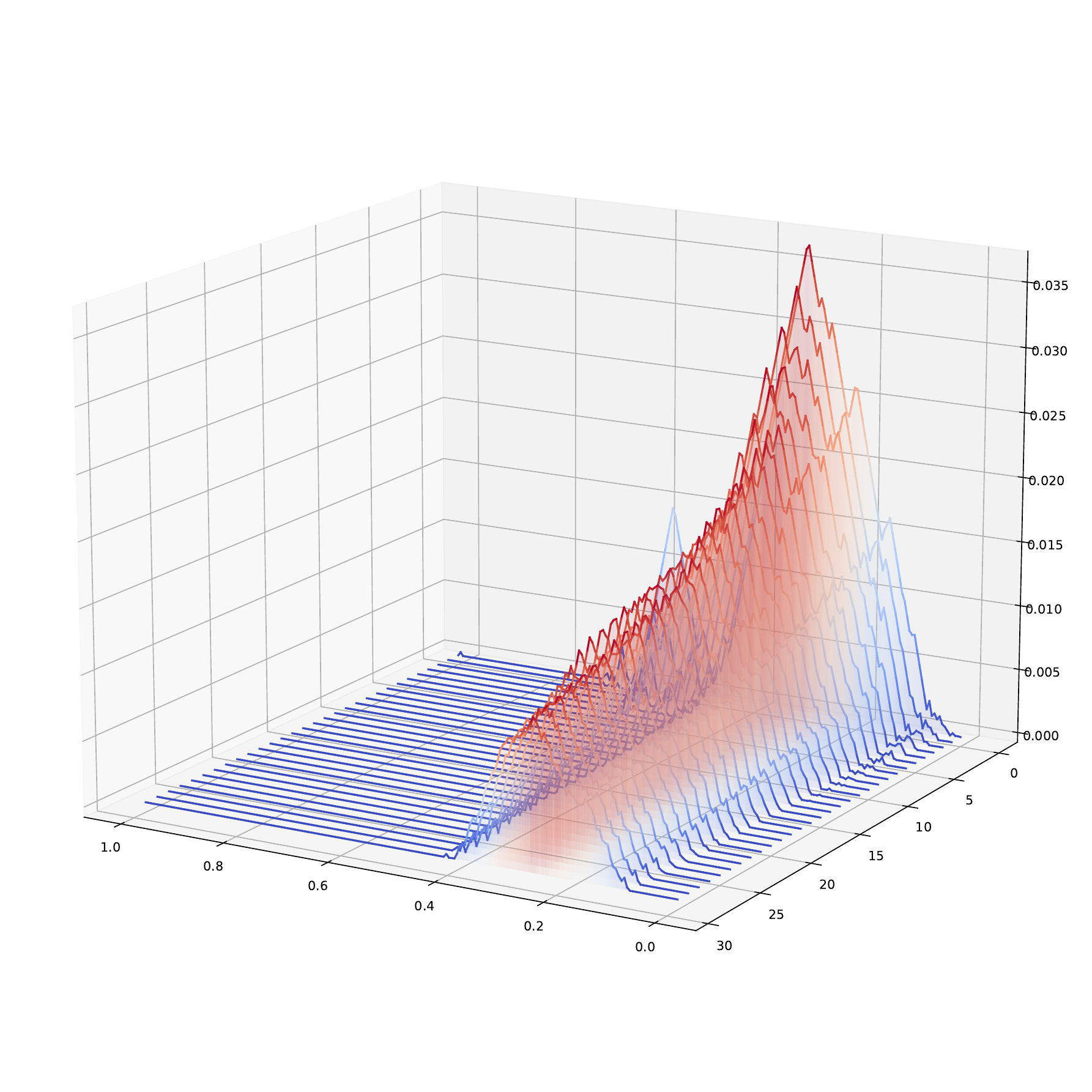}
        \subcaption{Persistence landscape}
        \label{fig:landscape2}
    \end{minipage}
    \caption{(\subref{fig:PD1})The persistence diagram of the CSI300 Index on June 29, 2020, illustrating the birth-death pairs of topological features represented as points; (\subref{fig:landscape2}) The corresponding $H_1$  persistence landscape.}

\end{figure}

Algorithm \ref{algo:landscape} is leveraged to obtain the persistence landscape from a given persistence diagram. Mathematically, for every birth-death pair \(p_i(a_i,b_i) \in D\), where \(D\) represents persistence diagrams, a piecewise linear function \(\Psi_p : \mathbb{R} \rightarrow [0,\infty]\) is defined as follows:
\begin{equation}\label{equ:piecewise}
\Psi_p(x) = \begin{cases}
x - a & \text{if } x \in \left[ a, \frac{a + b}{2} \right] \\
b - x & \text{if } x \in \left[ \frac{a + b}{2}, b \right] \\
0 & \text{if } x \notin \left[ a, b \right]
\end{cases}
\end{equation}
A persistence landscape (see \ref{fig:landscape2}) of the birth-death pairs \(p_i(a_i,b_i), i = 1, \dots,m\) is defined as a sequence of functions \(\lambda: \mathbb{N} \times \mathbb{R} \rightarrow [0,\infty]\), given by \(\lambda(k,x) = \lambda_k(x)\), where \(\lambda_k(x)\) denotes the \(k\)-th largest value of \(\{\Psi_{p_i}(x) \mid i = 1,\dots,m\}\). If the \(k\)-th largest value does not exist, then \(\lambda_k(x) = 0\). More specifically, \(\lambda: \mathbb{N} \times \mathbb{R} \rightarrow [0,\infty]\) is defined as follows:
\begin{equation}
    \lambda_k(x) = k\text{-}\max \{ \Psi_{(a_i, b_i)}(x) \mid (a_i, b_i) \in D \}
\end{equation}
Consequently, persistence landscapes form a subset of the Banach space \(L^p(\mathbb{N} \times \mathbb{R})\). The persistence landscape possesses a clear vector space structure (see Gidea and Katz\cite{gidea_topological_2018}), and it becomes a Banach space when endowed with the following norm:
\begin{equation}
    \|\lambda\|_p = \left( \sum_{k=1}^{\infty} \|\lambda_k\|_p^p \right)^{\frac{1}{p}},
\end{equation}
where \(\| \cdot \|\) denotes the \(L^p\)-norm, i.e., \(\|f\|_p = \left( \int_{\mathbb{R}} |f|^p \, dx \right)^{\frac{1}{p}}\), with integration taken with respect to the Lebesgue measure on \(\mathbb{R}\). Furthermore, Bubenik \cite{bubenik2015statistical} has demonstrated that the persistence landscape is stable with respect to the \(L^p\) norm for \(1 < p < \infty\). In this article, we computed the \(L^2\) norm for each dimension as a representation of the vectorized topological features. This is expressed as follows:
\begin{equation}
    F = [\|\lambda^{0}\|_2, \|\lambda^{1}\|_2, \dots, \|\lambda^{\text{dim}}\|_2] \in \mathbb{R}^{\text{dim}}
\end{equation}
where \(\text{dim}\) denotes the dimension of the persistence landscape. The vector \(F\) encapsulates the topological features, making it suitable for input into machine learning models.


\section{Methodology}\label{methodology}
\begin{figure}[H]
  \centering
  \includegraphics[width=1\textwidth]{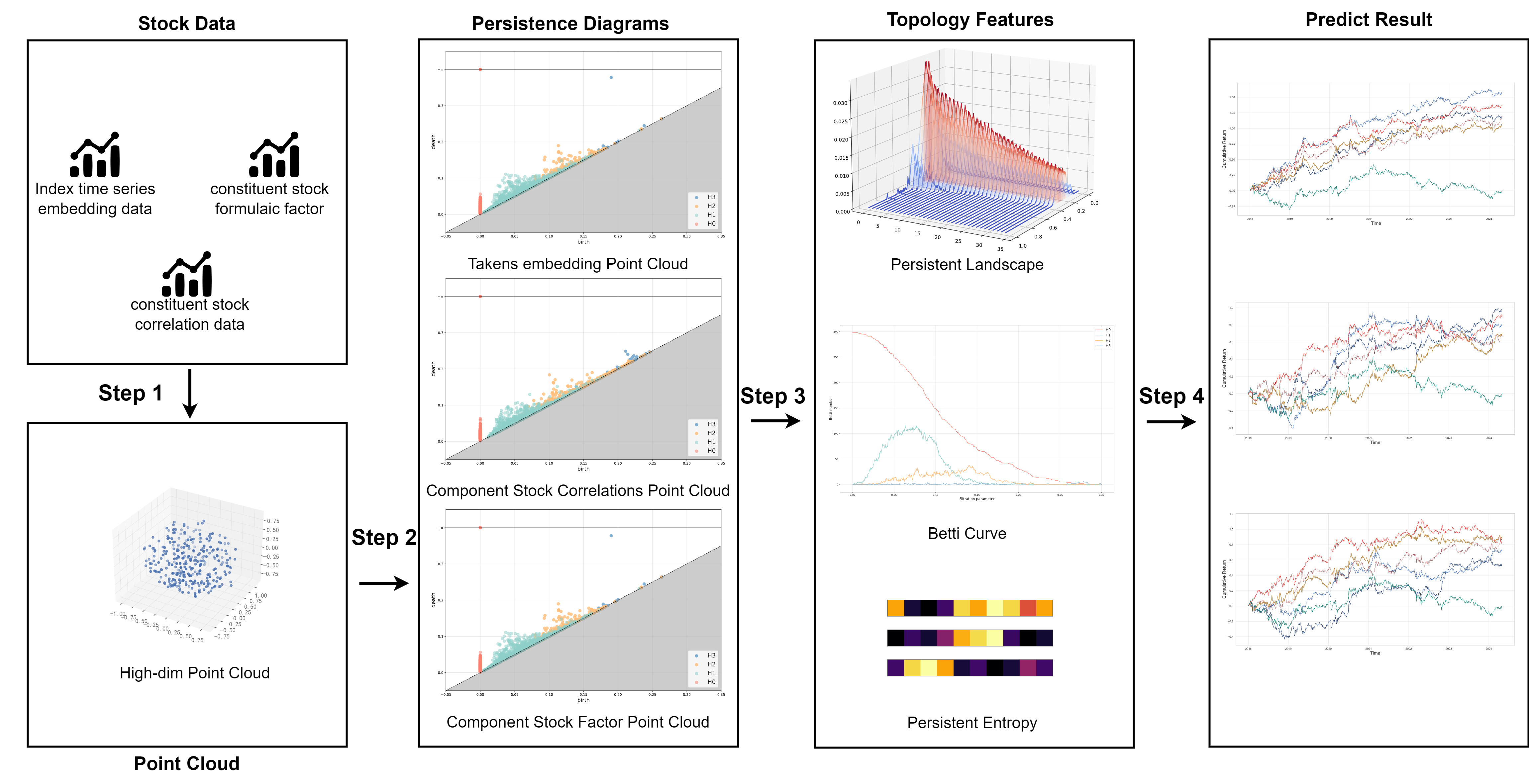}
  \caption{Workflow for Index Prediction. We build point clouds for each stock data set using three different ways(Section \ref{step1}). For each point cloud, we use persistent homology to get its persistence diagram(Section \ref{step 2}). We choose three topological features from each diagram(Section \ref{step3}), and use them as inputs in Section \ref{step4}.}
  \label{fig:fig1}
\end{figure}

In this section, we outline the main methodology of our study. The workflow is divided into four steps, as illustrated in figure \ref{fig:fig1}. The complete code for our framework is available at \url{https://github.com/Desman107/StockTDA}

\subsection{Step 1: Constructing High-Dimensional Point Clouds from Stock Data}\label{step1}

In Step 1, we outline the algorithm for constructing point clouds, which serves as the foundation for our TDA-based analysis. This study utilizes three types of point cloud constructions: Takens'
Embedding Point Cloud, Component Stock Correlation Point Cloud, and the Component Stock Formulaic Factor Point Cloud of index components. Detailed explanations for each construction method are provided below.

\subsubsection{Takens Embedding Point Cloud}\label{subsec_step2}

The construction of the Index Time Series Embedding Point Cloud leverages delay embedding techniques as discussed by Goel et al. \cite{goel_topological_2020}, which are effective for transforming historical stock return series into high-dimensional representations known as quasi-attractors. This method converts the time series observations \( \{x_0, x_1, x_2, \dots, x_t\} \) into phase space vectors \( Z = \{z_0, z_1, \dots, z_{t^\prime}\} \), where \( t^\prime = t - (p-1)\tau + 1 \), through delay embedding (Umeda\cite{umeda_time_2017},Tran and Hasegawa\cite{tran_topological_2019}). The delay vector is constructed as follows:

\begin{equation}
\begin{aligned}
z_k &= [x_k, x_{k+\tau}, \dots, x_{k+(p-1)\tau}] \in \mathbb{R}^p \\
    &= \begin{bmatrix}
x_1 & x_{1+\tau}  & \cdots & x_{1+(p-1)\tau} \\
x_2 & x_{2+\tau}  & \cdots & x_{2+(p-1)\tau} \\
\vdots & \vdots & \ddots & \vdots \\
x_{N-(p-1)\tau} & x_{N-(p-2)\tau}  & \cdots & x_N \\
\end{bmatrix}
\end{aligned}
\end{equation}

where \( \tau \) represents the lag and \( p \) is the embedding dimension. The quasi-attractor comprises the set of these delay vectors, capturing the system's behavior(Seversky et al. \cite{seversky_time-series_2016}) and preserving the topological properties of the original system (Takens\cite{rand_detecting_1981}, Packard et al.\cite{packard_geometry_1980}), making it suitable for persistent homology analysis.

In this article, the quasi-attractor is obtained as point-cloud data, facilitating topological data analysis. To capture fluctuations across different time scales—daily, weekly, monthly, and quarterly—we select lags that include the immediate return, along with 5, 20, and 60 trading day intervals (Jegadeesh and Titman\cite{jegadeesh2001profitability}). The quasi-attractor is then constructed using these selected lags from the index return rates, effectively representing multi-scale financial dynamics.

\subsubsection{Component Stock Correlations Point Cloud}\label{subsec_step3}
To analyze the complex correlations among stock constituents, we construct the Component Stock Correlations Point Cloud by transforming stock return data into a high-dimensional correlation matrix and subsequently into a low-dimensional spatial representation through Multidimensional Scaling (MDS)(Carroll and Arabie\cite{carroll1998multidimensional}). This approach was applied by Yen and Cheong\cite{yen_using_2021} and  Guo et al. \cite{guo_risk_2022}, which enables us to capture the relational structure within a given time window, thus forming a point cloud that encapsulates the temporal dependencies among stocks.

The process starts by selecting a sliding time window of $n$ days for each target date $t$. Let $r_{i,t}$ represent the return of stock $i$ on day $t$, and let $R_{i,t}$ and $R_{j,t}$ be the mean returns of stocks $i$ and $j$ over the selected window. We compute the correlation coefficient $C_{t}(i, j)$ between the time series of returns $R_{i,t}$ and $R_{j,t}$ as follows:

\[
C_{t}(i, j) = \frac{\sum_{k=1}^{l} (r_{i,t-k+1} - R_{i,t})(r_{j,t-k+1} - R_{j,t})}{\sqrt{\sum_{k=1}^{l} (r_{i,t-k+1} - R_{i,t})^2 \sum_{k=1}^{l} (r_{j,t-k+1} - R_{j,t})^2}}
\]

where $l$ is the length of the window. Once the correlation matrix $C_{t}$ is computed, we convert it into a distance matrix $D_{t}$ using the transformation introduced in Mantegna\cite{mantegna1999hierarchical}:

\[
D_{t}(i, j) = \sqrt{2(1 - C_{t}(i, j))}
\]

This distance matrix $D_{t}$ represents the dissimilarity between pairs of stocks, with smaller distances indicating stronger correlations. To generate the Component Stock Correlations Point Cloud in a lower-dimensional Euclidean space, we apply MDS to $D_{t}$, reducing the data to four dimensions while preserving the distance information. The resulting spatial coordinates form a point cloud for each time window, providing a representation suitable for topological analysis.



\subsubsection{Component Stock Factor Point Cloud}\label{subsec_step4}

The Component Stock Factor Point Cloud construction leverages structured factor data from the CSI300 constituent stocks on each target day \( t \). This approach builds upon previous researches, which demonstrated the efficacy of constituent stock data in market index prediction, as explored by Wang and Choi\cite{wang_market_2013} and Bareket and Pârv\cite{bareket_predicting_2024} . These studies highlighted how component-level data can be instrumental in capturing underlying market dynamics.

In our method, we utilize factor data generated from a deep reinforcement learning model as described by Yu et al.\cite{yu2023generating}. These formula-based factors, specifically tailored to reflect the unique features of the CSI300 constituents, are extracted for each trading day to capture temporal shifts in market-relevant characteristics. We then apply kernel Principal Component Analysis (kPCA), introduced by Schölkopf et al.\cite{scholkopf1998nonlinear}, with its application to TDA discussed by Reininghaus et al. \cite{reininghaus_stable_2015} and Kim et al. \cite{kim2018time}, to this high-dimensional Euclidean factor dataset to reduce dimensionality, effectively mapping the factor data onto a lower-dimensional space while retaining the core variability that represents essential market factors.

Mathematically, let \( F_{t} \in \mathbb{R}^{n \times d} \) denote the matrix of formula-based factors for the \( n \) constituent stocks on day \( t \), where \( d \) represents the number of factors extracted. Applying kPCA to \( F_{t} \) yields a transformed matrix \( F_{t}' \in \mathbb{R}^{n \times m} \) (with \( m < d \)), which serves as the embedded representation of the factor data. The rows of \( F_{t}' \) then constitute the points in our Component Stock Factor Point Cloud for day \( t \).


\subsection{Step 2: Constructing the Persistence Diagram Using Persistent Homology}\label{step 2}
In Step 2, we apply persistent homology to analyze the structural features of the high-dimensional point clouds generated in Step 1. The persistence diagram, capturing the birth and death of topological features across scales, effectively reveals the shape and connectivity of the financial data’s underlying geometry.

After evaluating various simplicial complexes, we selected the Alpha Complex (see Edelsbrunner et al. \cite{edelsbrunner2002topological}) for its balance of computational efficiency and feature extraction capability, critical given the large-scale complexity of financial market data. Compared to other constructions, such as Vietoris-Rips complexes and Čech complexes, the Alpha Complex is more computationally feasible while preserving key topological features, making it suitable for large datasets.

The Alpha Complex, constructed using the Gudhi library in Python, generates a simplicial complex from the point cloud metric, allowing efficient persistence calculations across dimensions (Maria et al.\cite{hong_gudhi_2014}). Gudhi optimizes Alpha Complex construction and persistence calculation, especially fitting for high-dimensional data in financial analysis. This method provides a persistence diagram capturing enduring topological features, revealing stable patterns and temporal structures within the CSI300 index components.

\subsection{Step 3: Extracting Topological Features from Persistence Diagram}\label{step3}

In Step 3, we extract three key topological features from the persistence diagrams generated in \ref{step 2}: the Betti curve, persistent entropy, and the \( L_2 \) norm of the persistence landscape. As introduced in \ref{sec_top_fea}, these features are derived from the persistence diagram and capture essential topological characteristics of the financial data, providing compact and informative representations for subsequent analysis.

\subsection{Step 4: Combining Topological Features and Model Evaluation}\label{step4}

In Step 4, we construct combinations of the multiple extracted vectorized topological features—Betti curve, persistent entropy, total persistence, and the \( L_2 \) norm of the persistence landscape,etc—by concatenating them into a single feature vector. These concatenated vectors are used as inputs for various common machine learning models, including XGBoost, LightGBM, RandomForest, SVM(rbf kernel), LSTM, MLP. By exploring all possible feature combinations, each concatenated vector serves as an input \( X \) for the classifiers, generating a diverse set of prediction outcomes. The results, which vary across different point cloud constructions, feature combinations, and models, are analyzed and compared in detail in Section \ref{Experiments}.



\section{Empirical Analysis and Results}\label{Experiments}

\subsection{Data}

The primary dataset used in this study consists of daily data from the CSI300 Index and its constituent stocks, spanning from January 2015 to June 2024. The CSI300 Index, which is a key benchmark for the Shanghai and Shenzhen stock markets, includes 300 of the largest and most representative stocks, providing a comprehensive view of market dynamics. The constituent data includes adjusted closing prices, returns, and a range of financial metrics, enabling a detailed analysis of both individual stock behavior and its aggregate impact on the index.

To ensure the robustness  of our empirical results, we extend our analysis to other major global indices, including the DAX30 Index, Hang Seng Index (HSI), and the UK's FTSE 100 Index. This broader dataset enhances the validity of our findings across different market conditions and regions.

The dataset is divided into:
- \textbf{Training Period (Jan 2015– Dec 2017)}: Used solely for model training to capture stable market patterns.
- \textbf{Prediction Period (Jan 2018– Jun 2024)}: For out-of-sample testing and evaluation of the \( t+1 \) daily return of the CSI300 Index.

To improve prediction over time, we apply a rolling prediction framework, as outlined by Clark and McCracken\cite{clark_improving_2009}. Each month, newly observed data from the prediction period is added to the training set, enabling the model to incorporate recent market changes and maintain robustness across the extended prediction period. For instance, when predicting stock movements for Jan 2021, the model is trained using data from Jan 2015 to Dec 2020. Subsequently, data from Jan 2021 is incorporated into the training set to prepare the model for predicting Feb 2021.

\subsection{Experiments}

In this setup, different combinations of topological features from persistence diagrams are evaluated as inputs (\( X \)) for binary classification models, predicting whether the \( t+1 \) daily return of the stock market Index will be positive. These topological features include the Betti curve, persistent entropy, total persistence and \( L_2 \) norm of the persistence landscape, as detailed in Section \ref{methodology}. Each feature combination provides a unique topological perspective on the market structure and serves as an input to models including XGBoost and other classifiers.

The experimental framework implements a rolling prediction method from 2018 onward. The model is initially trained on 2015–2017 data and retrained each month with the latest data from the prediction period. This approach allows dynamic adaptation to market changes. Model predictions and performance metrics, including accuracy, F1 score, cumulative return, and maximum drawdown, are recorded and analyzed in the Section \ref{Experiments} to assess the efficacy of each topological feature combination and point cloud construction method.



\begin{figure}[H]
\centering
\includegraphics[width=1\textwidth]{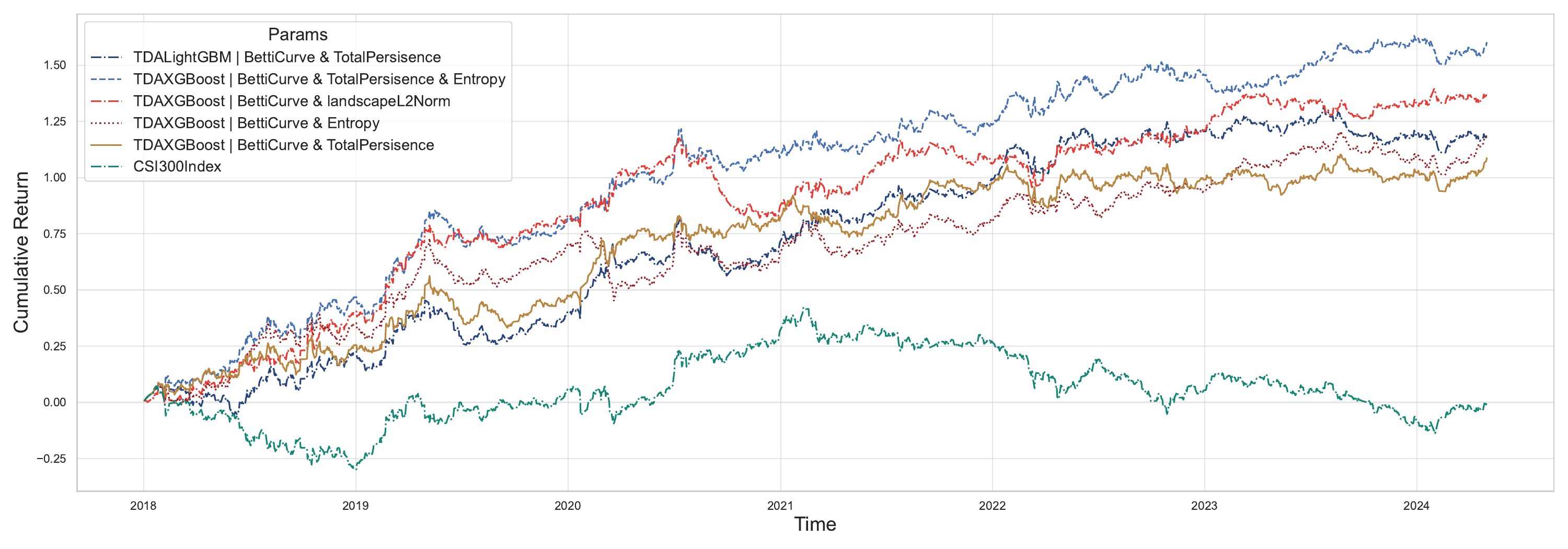}
\caption{Top 5 results on Takens embedding Point Cloud.}
\label{ReturnSeriesCloud}
\end{figure}

\begin{figure}[H]
\centering
\includegraphics[width=1\textwidth]{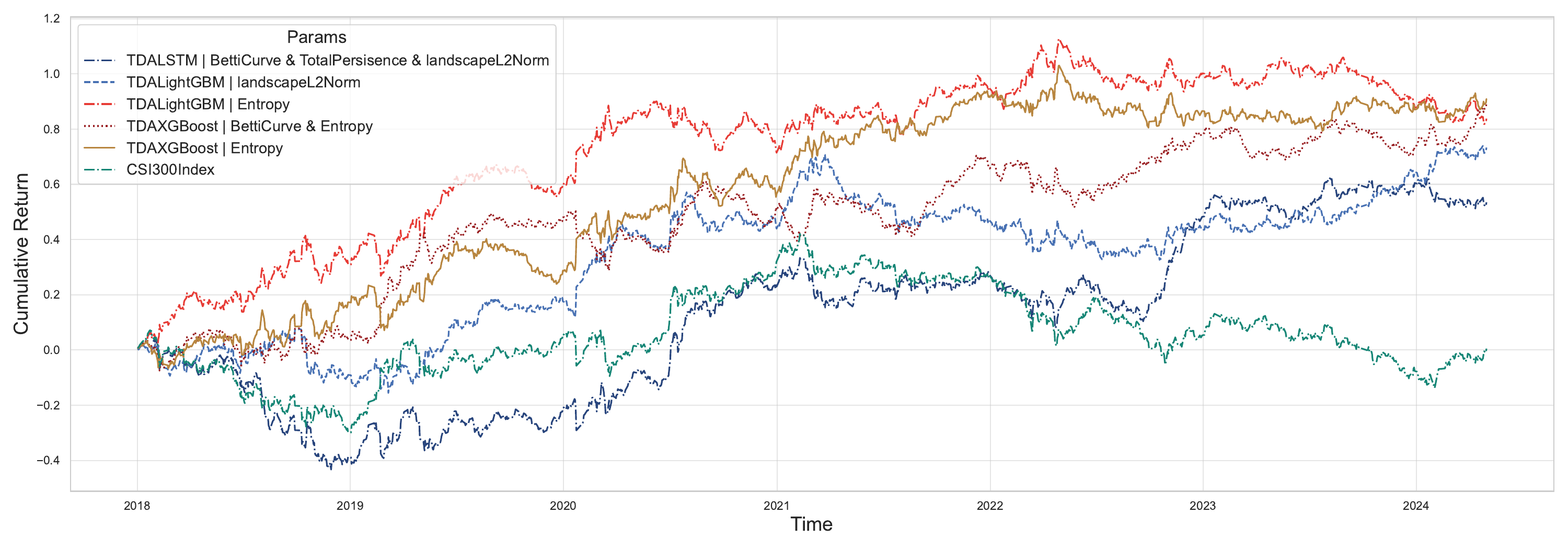}
\caption{Top 5 results on Component Stock Correlations Point Cloud. }
\label{CorrMDSCloud}
\end{figure}

\begin{figure}[H]
\centering
\includegraphics[width=1\textwidth]{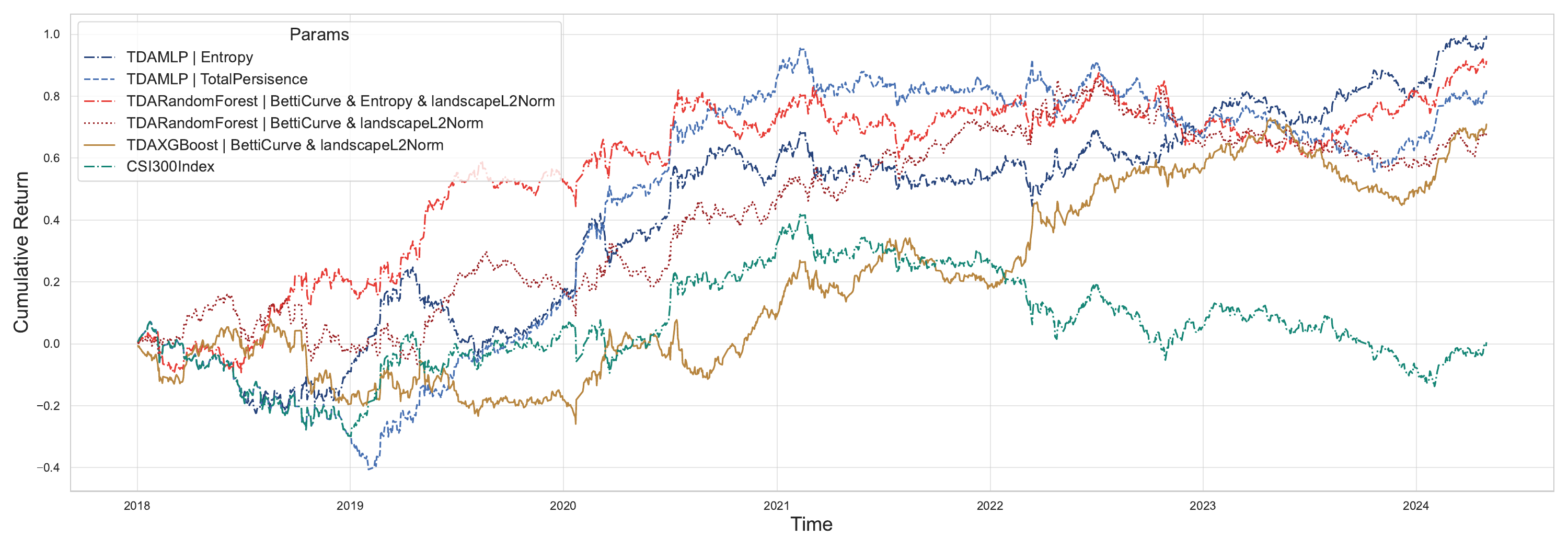}
\caption{Top 5 results on Component Stock Factor Point Cloud.}
\label{ConstituentCloud}
\end{figure}

\subsection{Empirical Results}
Overall, the classical time series attractor point cloud approach yields the best performance(see Figure \ref{ReturnSeriesCloud}), with the top five portfolios achieving cumulative returns exceeding 100\% during the period from 2018 to 2024. Specifically, combinations of topological features generated from the Takens embedding Point Cloud, including the Betti curve, total persistence, and persistent entropy, when used as inputs for the TDAXGBoost strategy, result in the highest cumulative return of over 150\%, alongside a relatively low maximum drawdown of 17.7\%. For the point cloud selection, the best way to Takens embedding point cloud achieve the highest average cumulative return which is 25.47\%, while constituent formulaic factor point cloud is -0.5\% and constituent formulaic correlation point cloud is -2.7\%.
\begin{figure}[H]
\centering
\includegraphics[width=1\textwidth]{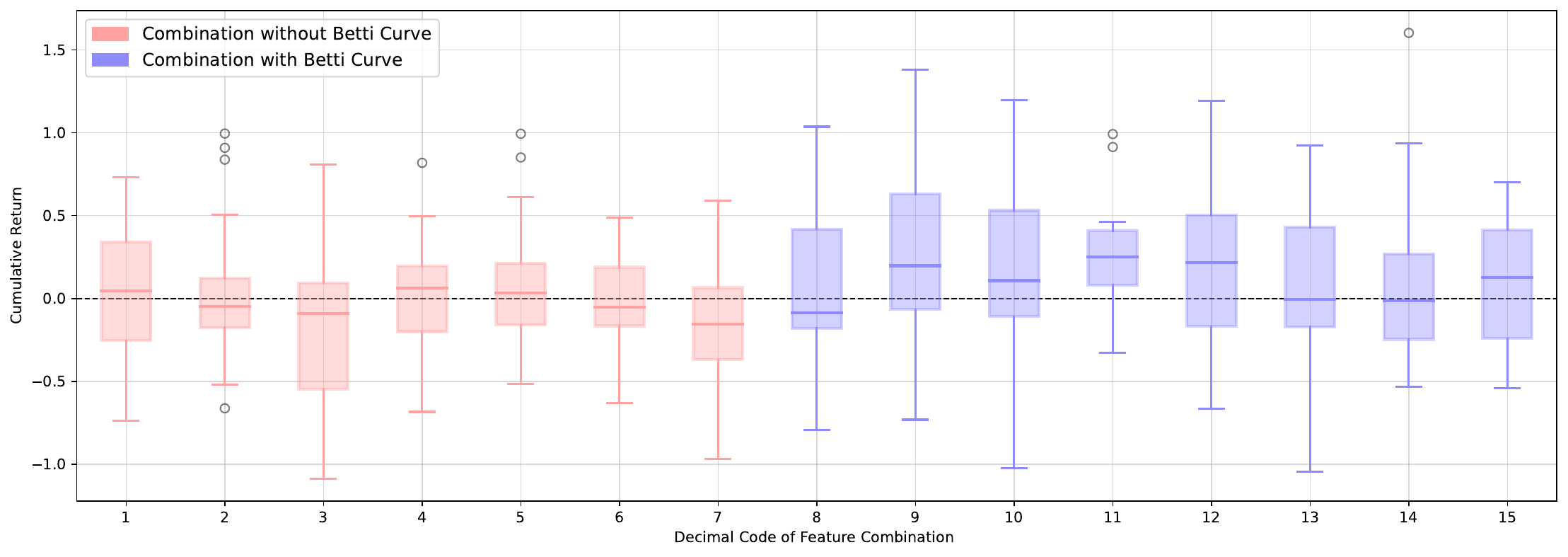}
\caption{Distribution of  cumulative returns observed in the experimental records across different topological feature combinations. The numbers on the x-axis correspond to specific feature combinations (refer to Table \ref{tab:topo_features}). Red boxes represent combinations without the Betti curve, while blue boxes indicate combinations that include the Betti curve.}
\label{Boxplot}
\end{figure}


Focusing solely on the impact of topological feature combinations, independent of point cloud structures and machine learning model categories, the combination of Betti curve, persistent entropy, and persistence landscape yields the highest median cumulative return of 25.22\%. Although the Betti curve alone does not achieve an ideal cumulative return (-8.54\%), combinations that include the Betti curve outperform those without it by approximately 13.31\%, as illustrated in Figure \ref{Boxplot}. A similar, though less pronounced, phenomenon is observed with other topological features, such as persistent entropy. Combinations that exclude persistent entropy yield a higher mean cumulative return (10.44\%) compared to those that include it (4.76\%).Both total persistence and persistence landscape exhibit a similar trend to persistent entropy, where combinations excluding these features perform better than those including them. However, this does not imply that these features are inherently "weak"; rather, when combined with the Betti curve, the overall performance improves significantly, highlighting their complementary value in enhancing model outcomes.

In terms of the backtest results over different periods, as shown in Figure \ref{ConstituentCloud}, the Random Forest model with Betti curves, persistence landscapes, and the persistence landscape $L_2$-norm as input yields the best returns before May 2020. However, after this period, it is replaced by the MLP model, which uses total persistence as input, and the MLP model, which takes persistent entropy as input, respectively. From Figures [\ref{ReturnSeriesCloud}- \ref{ConstituentCloud}], it is evident that for different point cloud construction methods, the corresponding optimal machine learning model and feature combinations vary in achieving the best returns. Notably, even for the same point cloud, the performance ranking of different feature combinations changes over time.

\begin{table}[H]
\centering
\caption{Empirical results of extended experiments on global stock market indices datasets. These results are based on Takens Embedding Point Cloud. The \texttt{FeaComb} column refers to the decimal codes listed in Table \ref{tab:topo_features}.}
\label{tab:global_result}
\begin{tabular}{lllllll}
\toprule[1.5pt]
Dataset & FeaComb & Model Name & Cum. Equity & Cum. Return & Max Drawdown & Accuracy \\
\midrule
DAX  & 9  & TDAMLP           & 3.4E+00 & 1.3E+00 & -2.9E-01 & 5.4E-01 \\
DAX  & 11 & TDASVM           & 2.1E+00 & 8.4E-01 & -3.9E-01 & 5.4E-01 \\
DAX  & 12 & TDALSTM          & 1.9E+00 & 7.6E-01 & -4.1E-01 & 5.4E-01 \\
DAX  & 3  & TDAMLP           & 1.9E+00 & 7.5E-01 & -3.9E-01 & 5.4E-01 \\
DAX  & 2  & TDALSTM          & 1.9E+00 & 7.4E-01 & -3.9E-01 & 5.4E-01 \\
FTSE & 6  & TDASVM           & 2.7E+00 & 1.1E+00 & -1.7E-01 & 5.4E-01 \\
FTSE & 6  & TDALightGBM      & 2.5E+00 & 1.0E+00 & -1.8E-01 & 5.2E-01 \\
FTSE & 14 & TDALSTM          & 2.2E+00 & 8.9E-01 & -2.0E-01 & 5.2E-01 \\
FTSE & 7  & TDALightGBM      & 1.9E+00 & 7.2E-01 & -2.2E-01 & 5.1E-01 \\
FTSE & 6  & TDAXGBoost       & 1.9E+00 & 7.2E-01 & -2.3E-01 & 5.1E-01 \\
HSI  & 15 & TDAXGBoost       & 3.1E+00 & 1.3E+00 & -2.4E-01 & 5.3E-01 \\
HSI  & 14 & TDAMLP           & 2.4E+00 & 1.0E+00 & -2.4E-01 & 5.3E-01 \\
HSI  & 11 & TDAXGBoost       & 2.2E+00 & 9.4E-01 & -4.2E-01 & 5.3E-01 \\
HSI  & 15 & TDARandomForest  & 2.2E+00 & 9.2E-01 & -2.5E-01 & 5.2E-01 \\
HSI  & 3  & TDALightGBM      & 1.9E+00 & 8.0E-01 & -2.6E-01 & 5.2E-01 \\
\bottomrule[1.5pt]
\end{tabular}
\end{table}

The extended experiments\footnote{Detailed results can be found at: \url{https://github.com/Desman107/StockTDA/blob/main/docs/empirical_result/global_result.md}} on global stock market indices (see Table \ref{tab:global_result}) further support our conclusions drawn from the CSI300 dataset. Leading TDA setups, particularly those incorporating the Betti curve, demonstrate superior predictive performance. Moreover, combining other topological features with the Betti curve significantly enhances prediction efficiency, reinforcing its pivotal role in effective TDA-based modeling.

\section{Conclusion}\label{conclusion}

In this paper, we investigate how different point cloud construction methods, combinations of topological features, and machine learning models impact stock index movement classification results. We break down the TDA-based stock movement classification process into four steps. First, we construct three different types of point clouds using both index price data and constituent stock data. Specifically, we create the Takens Embedding Point Cloud, the Component Stock Correlations Point Cloud, and the Component Stock Factor Point Cloud, each designed to capture the topological features of the index price, the correlations among index constituent stocks, and the stock factors, respectively.

Next, we compute persistent homology using alpha complexes and derive four common topological features, which are combined in 15 distinct configurations. These feature combinations are then concatenated into a single vector and input into six different machine learning models for classification.

Our empirical results show that, based on the Takens Embedding Point Cloud, the configuration involving topology features such as Betti curves, total persistence, and persistent entropy, when input into TDAXGBoost, yields the highest profit. However, not every configuration achieves this optimal outcome, highlighting the significance of our findings. Among the point cloud types, the Takens Embedding Point Cloud achieved the highest average cumulative return, followed by the Component Stock Factor Point Cloud and the Component Stock Correlations Point Cloud. In terms of topological feature combinations, those including the Betti curve consistently outperformed those without it, demonstrating its value. While other topological features should not be considered 'weak', combining them with the Betti curve significantly enhances classification accuracy compared to using Betti curves alone. Finally, in terms of machine learning models,  XGBoost and LightGBM demonstrated the highest accuracy and returns.

When classifying stock index movements, the relative ranking of optimal point cloud construction methods, feature combinations, and machine learning models shows slight variation over time. However, when the prediction target changes—for instance, shifting focus to stock market crashes—the optimal point cloud and feature combination also changes significantly. These findings highlight the dynamic nature of financial markets and the necessity of tailoring TDA configurations to specific prediction objectives and time frames.

We offer a new perspective for analyzing stock indices by considering the topological structure of constituent stock correlations and formulaic factors. While not all TDA configurations and models yield ideal performance, this research paves the way for further exploration to identify all potential methods for stock market analysis through Topological Data Analysis.

\section*{Acknowledgments}
The authors are thankful to Mr. Bogeng Huang, for valuable and professional suggestions inspired us and significantly promoted our research.

\bibliographystyle{unsrt}
\bibliography{templateArxiv.bbl}

\appendix
\section{Combination Code}

This section enumerates the topological feature combinations used in the study. Each combination is represented by a binary code, where a value of 1 indicates the inclusion of a specific feature, and a value of 0 indicates its exclusion. The binary code is then converted into a corresponding decimal code for easier reference. Table \ref{tab:topo_features} details the mapping between the binary and decimal codes for each topological feature combination.

\begin{table}[H]
\centering
\caption{Enumeration of Topological Feature Combinations}
\label{tab:topo_features}
\begin{tabular}{cccccc}
\toprule[1.5pt]
Betti Curve & Total Persistence & Entropy & Landscape $L_2$ Norm & Binary Code & Decimal Code \\
\midrule
0 & 0 & 0 & 1 & $(0001)_2$ & 1 \\
0 & 0 & 1 & 0 & $(0010)_2$ & 2 \\
0 & 0 & 1 & 1 & $(0011)_2$ & 3 \\
0 & 1 & 0 & 0 & $(0100)_2$ & 4 \\
0 & 1 & 0 & 1 & $(0101)_2$ & 5 \\
0 & 1 & 1 & 0 & $(0110)_2$ & 6 \\
0 & 1 & 1 & 1 & $(0111)_2$ & 7 \\
1 & 0 & 0 & 0 & $(1000)_2$ & 8 \\
1 & 0 & 0 & 1 & $(1001)_2$ & 9 \\
1 & 0 & 1 & 0 & $(1010)_2$ & 10 \\
1 & 0 & 1 & 1 & $(1011)_2$ & 11 \\
1 & 1 & 0 & 0 & $(1100)_2$ & 12 \\
1 & 1 & 0 & 1 & $(1101)_2$ & 13 \\
1 & 1 & 1 & 0 & $(1110)_2$ & 14 \\
1 & 1 & 1 & 1 & $(1111)_2$ & 15 \\
\bottomrule[1.5pt]
\end{tabular}
\end{table}

\section{Index explain}
The detail of our dataset description is show as follow:
\begin{itemize}
    \item \textbf{CSI300 (CSI)}: The CSI300 index is a key indicator of the Shanghai and Shenzhen stock markets, representing the performance of 300 of the largest and most liquid A-share stocks traded on these two exchanges. The dataset includes 294 constituents.

    \item \textbf{FTSE 100 (FTSE)}: The FTSE 100, also known as the Financial Times Stock Exchange 100 Index, is the leading stock market index in the United Kingdom. It tracks the performance of the 100 largest companies by market capitalization listed on the London Stock Exchange (LSE). The dataset contains 94 assets.

    \item \textbf{DAX 30 (DAX)}: The DAX 30 is a blue-chip stock market index consisting of 30 major German companies trading on the Frankfurt Stock Exchange. It reflects the performance of the largest and most liquid companies in Germany. The dataset contains 27 assets.

    \item \textbf{Hang Seng Index (HSI)}: The Hang Seng Index is a free float-adjusted market capitalization-weighted stock market index in Hong Kong, representing the performance of the 50 largest and most liquid companies. The dataset comprises 46 assets.
\end{itemize}

\newpage
\section{Empirical Results on CSI300}

This section presents the main metrics and parameters from the empirical results on the CSI300 dataset. Detailed results can be accessed at \url{https://github.com/Desman107/StockTDA/blob/main/docs/empirical_result/CSI300_result.md}. The \texttt{FeaComb} column in the following tables refers to the decimal codes listed in Table \ref{tab:topo_features}.

\begin{table}[H]
\begin{tabular}{lrlllll}
\toprule[1.5pt]
Cloud Type & FeaComb & Model Name & Cum. Equity & Cum. Return & Max Drawdown & Accuracy \\
\midrule
FactorCloud & 1 & TDALSTM & 7.41E-01 & -1.83E-01 & -4.02E-01 & 4.90E-01 \\
FactorCloud & 1 & TDALightGBM & 1.30E+00 & 3.76E-01 & -3.61E-01 & 5.01E-01 \\
FactorCloud & 1 & TDAMLP & 1.04E+00 & 1.53E-01 & -2.85E-01 & 5.00E-01 \\
FactorCloud & 1 & TDARandomForest & 6.85E-01 & -2.63E-01 & -4.90E-01 & 4.87E-01 \\
FactorCloud & 1 & TDASVM & 1.22E+00 & 3.15E-01 & -3.27E-01 & 5.05E-01 \\
FactorCloud & 1 & TDAXGBoost & 1.15E+00 & 2.54E-01 & -3.82E-01 & 5.04E-01 \\
FactorCloud & 2 & TDALSTM & 1.01E+00 & 1.28E-01 & -4.32E-01 & 5.01E-01 \\
FactorCloud & 2 & TDALightGBM & 7.92E-01 & -1.17E-01 & -3.79E-01 & 4.96E-01 \\
FactorCloud & 2 & TDAMLP & 2.41E+00 & 9.95E-01 & -2.63E-01 & 5.32E-01 \\
FactorCloud & 2 & TDARandomForest & 9.89E-01 & 1.04E-01 & -3.64E-01 & 5.00E-01 \\
FactorCloud & 2 & TDASVM & 7.15E-01 & -2.19E-01 & -4.34E-01 & 4.98E-01 \\
FactorCloud & 2 & TDAXGBoost & 4.59E-01 & -6.63E-01 & -6.00E-01 & 4.83E-01 \\
FactorCloud & 3 & TDALSTM & 8.95E-01 & 5.04E-03 & -4.53E-01 & 4.96E-01 \\
FactorCloud & 3 & TDALightGBM & 5.60E-01 & -4.63E-01 & -5.62E-01 & 4.79E-01 \\
FactorCloud & 3 & TDAMLP & 1.03E+00 & 1.50E-01 & -3.14E-01 & 4.96E-01 \\
FactorCloud & 3 & TDARandomForest & 7.87E-01 & -1.24E-01 & -3.28E-01 & 4.96E-01 \\
FactorCloud & 3 & TDASVM & 4.55E-01 & -6.72E-01 & -5.98E-01 & 4.83E-01 \\
FactorCloud & 3 & TDAXGBoost & 5.01E-01 & -5.74E-01 & -6.12E-01 & 4.85E-01 \\
FactorCloud & 4 & TDALSTM & 8.72E-01 & -2.04E-02 & -4.53E-01 & 4.97E-01 \\
FactorCloud & 4 & TDALightGBM & 4.49E-01 & -6.85E-01 & -6.37E-01 & 4.78E-01 \\
FactorCloud & 4 & TDAMLP & 2.02E+00 & 8.19E-01 & -3.94E-01 & 5.13E-01 \\
FactorCloud & 4 & TDARandomForest & 9.31E-01 & 4.44E-02 & -4.32E-01 & 5.02E-01 \\
FactorCloud & 4 & TDASVM & 8.72E-01 & -2.05E-02 & -4.14E-01 & 4.96E-01 \\
FactorCloud & 4 & TDAXGBoost & 5.67E-01 & -4.52E-01 & -5.28E-01 & 4.93E-01 \\
FactorCloud & 5 & TDALSTM & 9.15E-01 & 2.71E-02 & -4.35E-01 & 4.95E-01 \\
FactorCloud & 5 & TDALightGBM & 8.52E-01 & -4.39E-02 & -4.36E-01 & 5.00E-01 \\
FactorCloud & 5 & TDAMLP & 1.08E+00 & 1.95E-01 & -3.39E-01 & 4.94E-01 \\
FactorCloud & 5 & TDARandomForest & 5.88E-01 & -4.15E-01 & -5.28E-01 & 4.89E-01 \\
FactorCloud & 5 & TDASVM & 1.06E+00 & 1.72E-01 & -3.72E-01 & 5.06E-01 \\
FactorCloud & 5 & TDAXGBoost & 6.52E-01 & -3.11E-01 & -4.84E-01 & 4.89E-01 \\
FactorCloud & 6 & TDALSTM & 7.36E-01 & -1.91E-01 & -4.89E-01 & 4.91E-01 \\
FactorCloud & 6 & TDALightGBM & 8.12E-01 & -9.17E-02 & -3.79E-01 & 5.15E-01 \\
FactorCloud & 6 & TDAMLP & 1.05E+00 & 1.68E-01 & -4.65E-01 & 5.00E-01 \\
FactorCloud & 6 & TDARandomForest & 8.61E-01 & -3.36E-02 & -3.52E-01 & 5.06E-01 \\
FactorCloud & 6 & TDASVM & 1.19E+00 & 2.88E-01 & -3.79E-01 & 5.06E-01 \\
FactorCloud & 6 & TDAXGBoost & 8.13E-01 & -9.14E-02 & -4.77E-01 & 5.12E-01 \\
FactorCloud & 7 & TDALSTM & 7.03E-01 & -2.36E-01 & -5.30E-01 & 4.90E-01 \\
FactorCloud & 7 & TDALightGBM & 7.72E-01 & -1.43E-01 & -3.54E-01 & 4.96E-01 \\
FactorCloud & 7 & TDAMLP & 9.60E-01 & 7.48E-02 & -3.72E-01 & 4.97E-01 \\
FactorCloud & 7 & TDARandomForest & 1.61E+00 & 5.90E-01 & -3.52E-01 & 5.06E-01 \\
FactorCloud & 7 & TDASVM & 8.58E-01 & -3.65E-02 & -4.59E-01 & 5.09E-01 \\
FactorCloud & 7 & TDAXGBoost & 5.08E-01 & -5.62E-01 & -5.60E-01 & 4.78E-01 \\
\bottomrule[1.5pt]
\end{tabular}
\end{table}

\begin{tabular}{lrlllll}
\toprule[1.5pt]
Cloud Type & FeaComb & Model Name & Cum. Equity & Cum. Return & Max Drawdown & Accuracy \\
\midrule
FactorCloud & 8 & TDALSTM & 4.02E-01 & -7.95E-01 & -6.43E-01 & 4.89E-01 \\
FactorCloud & 8 & TDALightGBM & 1.53E+00 & 5.43E-01 & -4.59E-01 & 5.25E-01 \\
FactorCloud & 8 & TDAMLP & 1.13E+00 & 2.35E-01 & -4.06E-01 & 5.17E-01 \\
FactorCloud & 8 & TDARandomForest & 1.19E+00 & 2.90E-01 & -3.14E-01 & 5.11E-01 \\
FactorCloud & 8 & TDASVM & 6.75E-01 & -2.76E-01 & -3.90E-01 & 4.86E-01 \\
FactorCloud & 8 & TDAXGBoost & 7.85E-01 & -1.27E-01 & -5.04E-01 & 5.17E-01 \\
FactorCloud & 9 & TDALSTM & 1.29E+00 & 3.73E-01 & -3.73E-01 & 5.04E-01 \\
FactorCloud & 9 & TDALightGBM & 1.32E+00 & 3.95E-01 & -3.70E-01 & 5.22E-01 \\
FactorCloud & 9 & TDAMLP & 4.28E-01 & -7.32E-01 & -5.87E-01 & 4.76E-01 \\
FactorCloud & 9 & TDARandomForest & 1.77E+00 & 6.87E-01 & -2.69E-01 & 5.13E-01 \\
FactorCloud & 9 & TDASVM & 9.18E-01 & 2.99E-02 & -4.08E-01 & 4.98E-01 \\
FactorCloud & 9 & TDAXGBoost & 1.81E+00 & 7.10E-01 & -3.10E-01 & 5.25E-01 \\
FactorCloud & 10 & TDALSTM & 7.91E-01 & -1.18E-01 & -4.21E-01 & 5.00E-01 \\
FactorCloud & 10 & TDALightGBM & 1.22E+00 & 3.14E-01 & -2.90E-01 & 5.06E-01 \\
FactorCloud & 10 & TDAMLP & 3.20E-01 & -1.02E+00 & -6.94E-01 & 4.61E-01 \\
FactorCloud & 10 & TDARandomForest & 7.13E-01 & -2.22E-01 & -4.57E-01 & 5.08E-01 \\
FactorCloud & 10 & TDASVM & 8.92E-01 & 1.35E-03 & -3.20E-01 & 4.99E-01 \\
FactorCloud & 10 & TDAXGBoost & 7.47E-01 & -1.76E-01 & -4.39E-01 & 5.09E-01 \\
FactorCloud & 11 & TDALSTM & 9.86E-01 & 1.02E-01 & -3.77E-01 & 5.12E-01 \\
FactorCloud & 11 & TDALightGBM & 1.40E+00 & 4.55E-01 & -2.58E-01 & 5.14E-01 \\
FactorCloud & 11 & TDAMLP & 1.13E+00 & 2.39E-01 & -2.89E-01 & 5.02E-01 \\
FactorCloud & 11 & TDARandomForest & 2.22E+00 & 9.14E-01 & -2.50E-01 & 5.17E-01 \\
FactorCloud & 11 & TDASVM & 9.36E-01 & 5.01E-02 & -3.67E-01 & 4.94E-01 \\
FactorCloud & 11 & TDAXGBoost & 1.05E+00 & 1.66E-01 & -3.83E-01 & 5.24E-01 \\
FactorCloud & 12 & TDALSTM & 5.46E-01 & -4.89E-01 & -5.22E-01 & 4.98E-01 \\
FactorCloud & 12 & TDALightGBM & 1.37E+00 & 4.34E-01 & -3.66E-01 & 5.24E-01 \\
FactorCloud & 12 & TDAMLP & 6.24E-01 & -3.55E-01 & -4.09E-01 & 4.85E-01 \\
FactorCloud & 12 & TDARandomForest & 9.98E-01 & 1.14E-01 & -5.55E-01 & 5.02E-01 \\
FactorCloud & 12 & TDASVM & 1.18E+00 & 2.81E-01 & -3.40E-01 & 5.00E-01 \\
FactorCloud & 12 & TDAXGBoost & 1.62E+00 & 5.97E-01 & -4.08E-01 & 5.21E-01 \\
FactorCloud & 13 & TDALSTM & 5.74E-01 & -4.40E-01 & -4.65E-01 & 4.87E-01 \\
FactorCloud & 13 & TDALightGBM & 7.28E-01 & -2.02E-01 & -4.27E-01 & 5.13E-01 \\
FactorCloud & 13 & TDAMLP & 3.13E-01 & -1.05E+00 & -7.32E-01 & 4.90E-01 \\
FactorCloud & 13 & TDARandomForest & 1.64E+00 & 6.09E-01 & -4.05E-01 & 5.24E-01 \\
FactorCloud & 13 & TDASVM & 8.61E-01 & -3.39E-02 & -4.17E-01 & 4.97E-01 \\
FactorCloud & 13 & TDAXGBoost & 7.73E-01 & -1.42E-01 & -4.51E-01 & 5.18E-01 \\
FactorCloud & 14 & TDALSTM & 1.06E+00 & 1.76E-01 & -3.95E-01 & 5.06E-01 \\
FactorCloud & 14 & TDALightGBM & 5.94E-01 & -4.05E-01 & -5.55E-01 & 5.04E-01 \\
FactorCloud & 14 & TDAMLP & 1.37E+00 & 4.33E-01 & -3.66E-01 & 5.12E-01 \\
FactorCloud & 14 & TDARandomForest & 8.65E-01 & -2.94E-02 & -5.32E-01 & 5.07E-01 \\
FactorCloud & 14 & TDASVM & 7.11E-01 & -2.25E-01 & -4.86E-01 & 4.87E-01 \\
FactorCloud & 14 & TDAXGBoost & 8.05E-01 & -1.01E-01 & -4.57E-01 & 5.12E-01 \\
FactorCloud & 15 & TDALSTM & 1.08E+00 & 1.93E-01 & -2.99E-01 & 4.96E-01 \\
FactorCloud & 15 & TDALightGBM & 9.11E-01 & 2.25E-02 & -5.28E-01 & 5.11E-01 \\
FactorCloud & 15 & TDAMLP & 6.38E-01 & -3.34E-01 & -5.00E-01 & 4.86E-01 \\
FactorCloud & 15 & TDARandomForest & 1.62E+00 & 5.97E-01 & -3.88E-01 & 5.24E-01 \\
FactorCloud & 15 & TDASVM & 7.87E-01 & -1.23E-01 & -4.93E-01 & 4.98E-01 \\
FactorCloud & 15 & TDAXGBoost & 6.72E-01 & -2.81E-01 & -5.47E-01 & 5.15E-01 \\
\bottomrule[1.5pt]
\end{tabular}

\begin{tabular}{lrlllll}
\toprule[1.5pt]
Cloud Type & FeaComb & Model Name & Cum. Equity & Cum. Return & Max Drawdown & Accuracy \\
\midrule
CorrCloud & 1 & TDALSTM & 4.65E-01 & -6.50E-01 & -6.15E-01 & 4.85E-01 \\
CorrCloud & 1 & TDALightGBM & 1.85E+00 & 7.33E-01 & -3.34E-01 & 5.17E-01 \\
CorrCloud & 1 & TDAMLP & 8.40E-01 & -5.84E-02 & -5.43E-01 & 5.02E-01 \\
CorrCloud & 1 & TDARandomForest & 7.09E-01 & -2.28E-01 & -3.99E-01 & 5.00E-01 \\
CorrCloud & 1 & TDASVM & 4.25E-01 & -7.40E-01 & -6.26E-01 & 4.85E-01 \\
CorrCloud & 1 & TDAXGBoost & 1.29E+00 & 3.71E-01 & -4.90E-01 & 5.09E-01 \\
CorrCloud & 2 & TDALSTM & 9.33E-01 & 4.64E-02 & -4.29E-01 & 4.95E-01 \\
CorrCloud & 2 & TDALightGBM & 2.06E+00 & 8.38E-01 & -2.83E-01 & 5.19E-01 \\
CorrCloud & 2 & TDAMLP & 7.45E-01 & -1.78E-01 & -3.76E-01 & 5.06E-01 \\
CorrCloud & 2 & TDARandomForest & 1.48E+00 & 5.07E-01 & -2.93E-01 & 5.17E-01 \\
CorrCloud & 2 & TDASVM & 8.90E-01 & -3.21E-04 & -4.02E-01 & 4.95E-01 \\
CorrCloud & 2 & TDAXGBoost & 2.21E+00 & 9.09E-01 & -2.23E-01 & 5.21E-01 \\
CorrCloud & 3 & TDALSTM & 8.58E-01 & -3.73E-02 & -4.51E-01 & 4.99E-01 \\
CorrCloud & 3 & TDALightGBM & 9.95E-01 & 1.11E-01 & -3.61E-01 & 5.17E-01 \\
CorrCloud & 3 & TDAMLP & 1.16E+00 & 2.66E-01 & -2.95E-01 & 5.08E-01 \\
CorrCloud & 3 & TDARandomForest & 3.41E-01 & -9.59E-01 & -7.02E-01 & 5.09E-01 \\
CorrCloud & 3 & TDASVM & 4.79E-01 & -6.20E-01 & -6.11E-01 & 4.79E-01 \\
CorrCloud & 3 & TDAXGBoost & 8.38E-01 & -6.03E-02 & -3.46E-01 & 5.16E-01 \\
CorrCloud & 4 & TDALSTM & 9.89E-01 & 1.05E-01 & -4.23E-01 & 5.00E-01 \\
CorrCloud & 4 & TDALightGBM & 1.35E+00 & 4.13E-01 & -3.55E-01 & 5.06E-01 \\
CorrCloud & 4 & TDAMLP & 1.10E+00 & 2.12E-01 & -4.06E-01 & 5.07E-01 \\
CorrCloud & 4 & TDARandomForest & 9.67E-01 & 8.21E-02 & -3.60E-01 & 5.00E-01 \\
CorrCloud & 4 & TDASVM & 6.13E-01 & -3.73E-01 & -5.22E-01 & 4.94E-01 \\
CorrCloud & 4 & TDAXGBoost & 6.87E-01 & -2.60E-01 & -3.95E-01 & 4.92E-01 \\
CorrCloud & 5 & TDALSTM & 9.24E-01 & 3.74E-02 & -4.32E-01 & 4.99E-01 \\
CorrCloud & 5 & TDALightGBM & 7.63E-01 & -1.55E-01 & -4.89E-01 & 4.98E-01 \\
CorrCloud & 5 & TDAMLP & 8.15E-01 & -8.91E-02 & -5.30E-01 & 5.04E-01 \\
CorrCloud & 5 & TDARandomForest & 9.71E-01 & 8.64E-02 & -4.04E-01 & 5.11E-01 \\
CorrCloud & 5 & TDASVM & 5.32E-01 & -5.15E-01 & -5.49E-01 & 4.95E-01 \\
CorrCloud & 5 & TDAXGBoost & 7.56E-01 & -1.63E-01 & -5.18E-01 & 4.88E-01 \\
CorrCloud & 6 & TDALSTM & 8.60E-01 & -3.42E-02 & -4.25E-01 & 4.96E-01 \\
CorrCloud & 6 & TDALightGBM & 1.30E+00 & 3.75E-01 & -2.58E-01 & 5.06E-01 \\
CorrCloud & 6 & TDAMLP & 1.31E+00 & 3.87E-01 & -2.72E-01 & 5.00E-01 \\
CorrCloud & 6 & TDARandomForest & 1.45E+00 & 4.87E-01 & -2.90E-01 & 5.15E-01 \\
CorrCloud & 6 & TDASVM & 4.74E-01 & -6.31E-01 & -5.75E-01 & 4.89E-01 \\
CorrCloud & 6 & TDAXGBoost & 8.25E-01 & -7.73E-02 & -3.50E-01 & 4.92E-01 \\
CorrCloud & 7 & TDALSTM & 7.91E-01 & -1.18E-01 & -5.47E-01 & 4.84E-01 \\
CorrCloud & 7 & TDALightGBM & 5.98E-01 & -3.98E-01 & -4.45E-01 & 4.94E-01 \\
CorrCloud & 7 & TDAMLP & 6.69E-01 & -2.87E-01 & -5.23E-01 & 4.89E-01 \\
CorrCloud & 7 & TDARandomForest & 3.38E-01 & -9.69E-01 & -7.06E-01 & 4.89E-01 \\
CorrCloud & 7 & TDASVM & 3.38E-01 & -9.69E-01 & -6.92E-01 & 4.85E-01 \\
CorrCloud & 7 & TDAXGBoost & 6.71E-01 & -2.83E-01 & -4.42E-01 & 4.96E-01 \\
\bottomrule[1.5pt]
\end{tabular}

\begin{tabular}{lrlllll}
\toprule[1.5pt]
Cloud Type & FeaComb & Model Name & Cum. Equity & Cum. Return & Max Drawdown & Accuracy \\
\midrule
CorrCloud & 8 & TDALSTM & 1.51E+00 & 5.30E-01 & -3.07E-01 & 5.18E-01 \\
CorrCloud & 8 & TDALightGBM & 5.85E-01 & -4.20E-01 & -6.09E-01 & 4.96E-01 \\
CorrCloud & 8 & TDAMLP & 7.92E-01 & -1.18E-01 & -5.37E-01 & 4.95E-01 \\
CorrCloud & 8 & TDARandomForest & 9.16E-01 & 2.78E-02 & -3.44E-01 & 5.00E-01 \\
CorrCloud & 8 & TDASVM & 8.41E-01 & -5.71E-02 & -4.57E-01 & 4.99E-01 \\
CorrCloud & 8 & TDAXGBoost & 5.48E-01 & -4.85E-01 & -5.46E-01 & 4.96E-01 \\
CorrCloud & 9 & TDALSTM & 1.41E+00 & 4.58E-01 & -4.26E-01 & 4.98E-01 \\
CorrCloud & 9 & TDALightGBM & 9.14E-01 & 2.64E-02 & -3.40E-01 & 5.02E-01 \\
CorrCloud & 9 & TDAMLP & 7.68E-01 & -1.48E-01 & -5.51E-01 & 5.07E-01 \\
CorrCloud & 9 & TDARandomForest & 9.69E-01 & 8.49E-02 & -3.54E-01 & 5.19E-01 \\
CorrCloud & 9 & TDASVM & 9.20E-01 & 3.29E-02 & -3.57E-01 & 5.02E-01 \\
CorrCloud & 9 & TDAXGBoost & 5.67E-01 & -4.51E-01 & -5.36E-01 & 4.91E-01 \\
CorrCloud & 10 & TDALSTM & 7.39E-01 & -1.87E-01 & -4.14E-01 & 4.96E-01 \\
CorrCloud & 10 & TDALightGBM & 1.10E+00 & 2.13E-01 & -4.06E-01 & 5.06E-01 \\
CorrCloud & 10 & TDAMLP & 1.16E+00 & 2.62E-01 & -3.10E-01 & 5.09E-01 \\
CorrCloud & 10 & TDARandomForest & 8.19E-01 & -8.35E-02 & -4.39E-01 & 5.17E-01 \\
CorrCloud & 10 & TDASVM & 8.37E-01 & -6.20E-02 & -5.14E-01 & 4.96E-01 \\
CorrCloud & 10 & TDAXGBoost & 2.17E+00 & 8.92E-01 & -2.05E-01 & 5.26E-01 \\
CorrCloud & 11 & TDALSTM & 1.25E+00 & 3.41E-01 & -3.35E-01 & 5.02E-01 \\
CorrCloud & 11 & TDALightGBM & 1.19E+00 & 2.93E-01 & -3.47E-01 & 5.11E-01 \\
CorrCloud & 11 & TDAMLP & 1.23E+00 & 3.20E-01 & -3.00E-01 & 5.04E-01 \\
CorrCloud & 11 & TDARandomForest & 1.37E+00 & 4.31E-01 & -2.62E-01 & 5.26E-01 \\
CorrCloud & 11 & TDASVM & 7.44E-01 & -1.80E-01 & -5.03E-01 & 4.93E-01 \\
CorrCloud & 11 & TDAXGBoost & 9.57E-01 & 7.23E-02 & -5.29E-01 & 4.99E-01 \\
CorrCloud & 12 & TDALSTM & 1.11E+00 & 2.21E-01 & -4.82E-01 & 5.04E-01 \\
CorrCloud & 12 & TDALightGBM & 4.58E-01 & -6.66E-01 & -5.99E-01 & 4.97E-01 \\
CorrCloud & 12 & TDAMLP & 6.84E-01 & -2.63E-01 & -3.59E-01 & 4.99E-01 \\
CorrCloud & 12 & TDARandomForest & 1.10E+00 & 2.10E-01 & -4.14E-01 & 4.98E-01 \\
CorrCloud & 12 & TDASVM & 9.98E-01 & 1.14E-01 & -3.40E-01 & 5.00E-01 \\
CorrCloud & 12 & TDAXGBoost & 6.27E-01 & -3.52E-01 & -5.46E-01 & 4.89E-01 \\
CorrCloud & 13 & TDALSTM & 1.52E+00 & 5.35E-01 & -4.04E-01 & 5.22E-01 \\
CorrCloud & 13 & TDALightGBM & 6.02E-01 & -3.92E-01 & -5.26E-01 & 5.06E-01 \\
CorrCloud & 13 & TDAMLP & 7.46E-01 & -1.78E-01 & -5.05E-01 & 4.91E-01 \\
CorrCloud & 13 & TDARandomForest & 1.41E+00 & 4.59E-01 & -2.70E-01 & 5.06E-01 \\
CorrCloud & 13 & TDASVM & 7.62E-01 & -1.56E-01 & -4.88E-01 & 4.98E-01 \\
CorrCloud & 13 & TDAXGBoost & 1.25E+00 & 3.40E-01 & -3.36E-01 & 5.06E-01 \\
CorrCloud & 14 & TDALSTM & 6.54E-01 & -3.08E-01 & -5.10E-01 & 4.91E-01 \\
CorrCloud & 14 & TDALightGBM & 5.23E-01 & -5.33E-01 & -5.43E-01 & 4.90E-01 \\
CorrCloud & 14 & TDAMLP & 8.26E-01 & -7.52E-02 & -3.63E-01 & 5.03E-01 \\
CorrCloud & 14 & TDARandomForest & 1.20E+00 & 2.94E-01 & -3.36E-01 & 5.01E-01 \\
CorrCloud & 14 & TDASVM & 8.91E-01 & 7.34E-04 & -3.36E-01 & 4.93E-01 \\
CorrCloud & 14 & TDAXGBoost & 6.91E-01 & -2.54E-01 & -4.46E-01 & 5.03E-01 \\
CorrCloud & 15 & TDALSTM & 8.29E-01 & -7.19E-02 & -4.38E-01 & 5.07E-01 \\
CorrCloud & 15 & TDALightGBM & 5.18E-01 & -5.42E-01 & -6.00E-01 & 4.96E-01 \\
CorrCloud & 15 & TDAMLP & 1.13E+00 & 2.36E-01 & -4.50E-01 & 5.16E-01 \\
CorrCloud & 15 & TDARandomForest & 1.37E+00 & 4.31E-01 & -3.08E-01 & 5.18E-01 \\
CorrCloud & 15 & TDASVM & 9.47E-01 & 6.14E-02 & -3.82E-01 & 4.99E-01 \\
CorrCloud & 15 & TDAXGBoost & 5.68E-01 & -4.50E-01 & -4.99E-01 & 4.95E-01 \\
\bottomrule[1.5pt]
\end{tabular}

\begin{tabular}{lrlllll}
\toprule[1.5pt]
Cloud Type & FeaComb & Model Name & Cum. Equity & Cum. Return & Max Drawdown & Accuracy \\
\midrule
Takens embedding & 1 & TDALSTM & 1.09E+00 & 2.04E-01 & -3.74E-01 & 5.06E-01 \\
Takens embedding & 1 & TDALightGBM & 1.26E+00 & 3.50E-01 & -4.96E-01 & 5.02E-01 \\
Takens embedding & 1 & TDAMLP & 5.46E-01 & -4.88E-01 & -4.91E-01 & 4.90E-01 \\
Takens embedding & 1 & TDARandomForest & 8.37E-01 & -6.25E-02 & -4.05E-01 & 4.89E-01 \\
Takens embedding & 1 & TDASVM & 4.52E-01 & -6.79E-01 & -6.01E-01 & 4.77E-01 \\
Takens embedding & 1 & TDAXGBoost & 1.41E+00 & 4.59E-01 & -4.44E-01 & 5.13E-01 \\
Takens embedding & 2 & TDALSTM & 8.97E-01 & 7.73E-03 & -4.00E-01 & 4.95E-01 \\
Takens embedding & 2 & TDALightGBM & 7.69E-01 & -1.47E-01 & -5.19E-01 & 4.96E-01 \\
Takens embedding & 2 & TDAMLP & 7.50E-01 & -1.72E-01 & -4.94E-01 & 5.02E-01 \\
Takens embedding & 2 & TDARandomForest & 6.57E-01 & -3.04E-01 & -5.29E-01 & 4.97E-01 \\
Takens embedding & 2 & TDASVM & 8.08E-01 & -9.69E-02 & -4.26E-01 & 5.16E-01 \\
Takens embedding & 2 & TDAXGBoost & 5.30E-01 & -5.19E-01 & -5.79E-01 & 5.00E-01 \\
Takens embedding & 3 & TDALSTM & 7.58E-01 & -1.61E-01 & -4.60E-01 & 4.96E-01 \\
Takens embedding & 3 & TDALightGBM & 6.94E-01 & -2.50E-01 & -4.78E-01 & 4.83E-01 \\
Takens embedding & 3 & TDAMLP & 2.00E+00 & 8.10E-01 & -3.33E-01 & 5.21E-01 \\
Takens embedding & 3 & TDARandomForest & 3.00E-01 & -1.09E+00 & -7.04E-01 & 4.72E-01 \\
Takens embedding & 3 & TDASVM & 1.47E+00 & 5.00E-01 & -3.11E-01 & 5.22E-01 \\
Takens embedding & 3 & TDAXGBoost & 9.30E-01 & 4.28E-02 & -3.37E-01 & 4.92E-01 \\
Takens embedding & 4 & TDALSTM & 1.03E+00 & 1.44E-01 & -3.99E-01 & 5.05E-01 \\
Takens embedding & 4 & TDALightGBM & 6.20E-01 & -3.63E-01 & -5.44E-01 & 4.96E-01 \\
Takens embedding & 4 & TDAMLP & 8.94E-01 & 3.45E-03 & -3.73E-01 & 4.99E-01 \\
Takens embedding & 4 & TDARandomForest & 1.04E+00 & 1.51E-01 & -4.20E-01 & 5.13E-01 \\
Takens embedding & 4 & TDASVM & 1.19E+00 & 2.94E-01 & -3.41E-01 & 5.08E-01 \\
Takens embedding & 4 & TDAXGBoost & 1.46E+00 & 4.95E-01 & -3.12E-01 & 5.15E-01 \\
Takens embedding & 5 & TDALSTM & 7.11E-01 & -2.25E-01 & -4.59E-01 & 4.98E-01 \\
Takens embedding & 5 & TDALightGBM & 1.64E+00 & 6.11E-01 & -3.24E-01 & 5.16E-01 \\
Takens embedding & 5 & TDAMLP & 2.41E+00 & 9.94E-01 & -1.96E-01 & 5.18E-01 \\
Takens embedding & 5 & TDARandomForest & 1.11E+00 & 2.20E-01 & -3.35E-01 & 5.04E-01 \\
Takens embedding & 5 & TDASVM & 1.46E+00 & 4.97E-01 & -3.20E-01 & 5.07E-01 \\
Takens embedding & 5 & TDAXGBoost & 2.09E+00 & 8.51E-01 & -3.38E-01 & 5.21E-01 \\
Takens embedding & 6 & TDALSTM & 4.84E-01 & -6.10E-01 & -5.72E-01 & 4.87E-01 \\
Takens embedding & 6 & TDALightGBM & 8.93E-01 & 3.03E-03 & -3.29E-01 & 4.97E-01 \\
Takens embedding & 6 & TDAMLP & 8.30E-01 & -7.04E-02 & -4.59E-01 & 5.07E-01 \\
Takens embedding & 6 & TDARandomForest & 5.80E-01 & -4.29E-01 & -5.53E-01 & 4.91E-01 \\
Takens embedding & 6 & TDASVM & 5.53E-01 & -4.76E-01 & -5.78E-01 & 4.96E-01 \\
Takens embedding & 6 & TDAXGBoost & 1.08E+00 & 1.95E-01 & -3.04E-01 & 5.04E-01 \\
Takens embedding & 7 & TDALSTM & 7.51E-01 & -1.71E-01 & -4.33E-01 & 5.05E-01 \\
Takens embedding & 7 & TDALightGBM & 1.38E+00 & 4.40E-01 & -3.24E-01 & 5.02E-01 \\
Takens embedding & 7 & TDAMLP & 9.22E-01 & 3.43E-02 & -4.56E-01 & 5.02E-01 \\
Takens embedding & 7 & TDARandomForest & 5.67E-01 & -4.51E-01 & -5.35E-01 & 4.96E-01 \\
Takens embedding & 7 & TDASVM & 1.06E+00 & 1.70E-01 & -4.09E-01 & 5.18E-01 \\
Takens embedding & 7 & TDAXGBoost & 1.07E+00 & 1.86E-01 & -2.75E-01 & 5.00E-01 \\
\bottomrule[1.5pt]
\end{tabular}

\begin{tabular}{lrlllll}
\toprule[1.5pt]
Cloud Type & FeaComb & Model Name & Cum. Equity & Cum. Return & Max Drawdown & Accuracy \\
\midrule
Takens embedding & 8 & TDALSTM & 1.41E+00 & 4.60E-01 & -3.24E-01 & 5.13E-01 \\
Takens embedding & 8 & TDALightGBM & 1.84E+00 & 7.26E-01 & -3.10E-01 & 5.14E-01 \\
Takens embedding & 8 & TDAMLP & 7.95E-01 & -1.14E-01 & -4.19E-01 & 4.95E-01 \\
Takens embedding & 8 & TDARandomForest & 7.73E-01 & -1.42E-01 & -5.35E-01 & 4.85E-01 \\
Takens embedding & 8 & TDASVM & 7.35E-01 & -1.92E-01 & -4.91E-01 & 4.94E-01 \\
Takens embedding & 8 & TDAXGBoost & 2.51E+00 & 1.04E+00 & -2.15E-01 & 5.16E-01 \\
Takens embedding & 9 & TDALSTM & 6.21E-01 & -3.60E-01 & -5.16E-01 & 5.04E-01 \\
Takens embedding & 9 & TDALightGBM & 2.52E+00 & 1.04E+00 & -3.12E-01 & 5.11E-01 \\
Takens embedding & 9 & TDAMLP & 2.05E+00 & 8.35E-01 & -3.90E-01 & 5.16E-01 \\
Takens embedding & 9 & TDARandomForest & 1.21E+00 & 3.09E-01 & -4.08E-01 & 5.14E-01 \\
Takens embedding & 9 & TDASVM & 8.08E-01 & -9.75E-02 & -5.16E-01 & 4.89E-01 \\
Takens embedding & 9 & TDAXGBoost & 3.54E+00 & 1.38E+00 & -3.05E-01 & 5.21E-01 \\
Takens embedding & 10 & TDALSTM & 1.43E+00 & 4.73E-01 & -4.69E-01 & 5.16E-01 \\
Takens embedding & 10 & TDALightGBM & 2.01E+00 & 8.15E-01 & -2.78E-01 & 5.18E-01 \\
Takens embedding & 10 & TDAMLP & 2.39E+00 & 9.87E-01 & -3.34E-01 & 5.18E-01 \\
Takens embedding & 10 & TDARandomForest & 1.54E+00 & 5.50E-01 & -2.69E-01 & 5.04E-01 \\
Takens embedding & 10 & TDASVM & 8.87E-01 & -4.20E-03 & -4.46E-01 & 4.97E-01 \\
Takens embedding & 10 & TDAXGBoost & 2.95E+00 & 1.20E+00 & -2.75E-01 & 5.28E-01 \\
Takens embedding & 11 & TDALSTM & 1.16E+00 & 2.65E-01 & -3.13E-01 & 5.09E-01 \\
Takens embedding & 11 & TDALightGBM & 1.03E+00 & 1.49E-01 & -4.28E-01 & 5.06E-01 \\
Takens embedding & 11 & TDAMLP & 6.42E-01 & -3.28E-01 & -4.96E-01 & 4.93E-01 \\
Takens embedding & 11 & TDARandomForest & 1.41E+00 & 4.61E-01 & -2.66E-01 & 5.06E-01 \\
Takens embedding & 11 & TDASVM & 9.47E-01 & 6.15E-02 & -4.48E-01 & 5.00E-01 \\
Takens embedding & 11 & TDAXGBoost & 2.40E+00 & 9.92E-01 & -2.15E-01 & 5.17E-01 \\
Takens embedding & 12 & TDALSTM & 1.12E+00 & 2.27E-01 & -4.21E-01 & 5.25E-01 \\
Takens embedding & 12 & TDALightGBM & 2.94E+00 & 1.19E+00 & -2.30E-01 & 5.26E-01 \\
Takens embedding & 12 & TDAMLP & 2.08E+00 & 8.50E-01 & -3.22E-01 & 5.17E-01 \\
Takens embedding & 12 & TDARandomForest & 1.51E+00 & 5.25E-01 & -3.99E-01 & 5.07E-01 \\
Takens embedding & 12 & TDASVM & 1.01E+00 & 1.25E-01 & -4.42E-01 & 4.98E-01 \\
Takens embedding & 12 & TDAXGBoost & 2.64E+00 & 1.09E+00 & -2.12E-01 & 5.14E-01 \\
Takens embedding & 13 & TDALSTM & 1.07E+00 & 1.83E-01 & -3.97E-01 & 4.98E-01 \\
Takens embedding & 13 & TDALightGBM & 1.82E+00 & 7.16E-01 & -2.53E-01 & 5.15E-01 \\
Takens embedding & 13 & TDAMLP & 7.87E-01 & -1.23E-01 & -4.19E-01 & 5.11E-01 \\
Takens embedding & 13 & TDARandomForest & 2.24E+00 & 9.22E-01 & -3.04E-01 & 5.07E-01 \\
Takens embedding & 13 & TDASVM & 9.12E-01 & 2.40E-02 & -4.46E-01 & 4.94E-01 \\
Takens embedding & 13 & TDAXGBoost & 1.22E+00 & 3.14E-01 & -3.60E-01 & 5.08E-01 \\
Takens embedding & 14 & TDALSTM & 9.20E-01 & 3.27E-02 & -4.64E-01 & 5.07E-01 \\
Takens embedding & 14 & TDALightGBM & 2.27E+00 & 9.37E-01 & -3.05E-01 & 5.33E-01 \\
Takens embedding & 14 & TDAMLP & 6.51E-01 & -3.14E-01 & -4.21E-01 & 4.91E-01 \\
Takens embedding & 14 & TDARandomForest & 2.08E+00 & 8.46E-01 & -4.25E-01 & 5.12E-01 \\
Takens embedding & 14 & TDASVM & 1.08E+00 & 1.90E-01 & -3.91E-01 & 5.01E-01 \\
Takens embedding & 14 & TDAXGBoost & 4.43E+00 & 1.60E+00 & -1.77E-01 & 5.34E-01 \\
Takens embedding & 15 & TDALSTM & 1.28E+00 & 3.61E-01 & -4.28E-01 & 5.09E-01 \\
Takens embedding & 15 & TDALightGBM & 1.80E+00 & 7.01E-01 & -2.70E-01 & 5.16E-01 \\
Takens embedding & 15 & TDAMLP & 6.43E-01 & -3.25E-01 & -4.27E-01 & 4.87E-01 \\
Takens embedding & 15 & TDARandomForest & 1.54E+00 & 5.46E-01 & -3.56E-01 & 5.15E-01 \\
Takens embedding & 15 & TDASVM & 1.21E+00 & 3.07E-01 & -4.07E-01 & 5.02E-01 \\
Takens embedding & 15 & TDAXGBoost & 1.63E+00 & 6.02E-01 & -3.66E-01 & 5.18E-01 \\
\bottomrule[1.5pt]
\end{tabular}

\end{document}